\documentclass[runningheads]{llncs}

\usepackage[mobile]{eccv}

\usepackage{eccvabbrv}
 
\usepackage{graphicx}
\usepackage{booktabs}
\usepackage[pagebackref,breaklinks,colorlinks]{hyperref}
\usepackage[capitalize]{cleveref}
\usepackage{relsize}

\usepackage{placeins} 
\usepackage{pifont}
\usepackage[table]{xcolor} 
\usepackage{caption}
\usepackage{booktabs}
\usepackage{multirow}
\usepackage[accsupp]{axessibility}  

\usepackage{hyperref}

\usepackage{orcidlink}

\begin{document}

\title{Generative Refocusing:\\Flexible Defocus Control from a Single Image} 

\titlerunning{Generative Refocusing}

\author{Chun-Wei Tuan Mu\inst{1} \and Cheng-De Fan\inst{1} \and Jia-Bin Huang\inst{2} \and Yu-Lun Liu\inst{1}}

\authorrunning{C.-W.~Tuan Mu et al.}

\institute{\textsuperscript{\rm 1} National Yang Ming Chiao Tung University, \textsuperscript{\rm 2} University of Maryland, College Park}

\maketitle

\begin{center}
  \includegraphics[width=\textwidth]{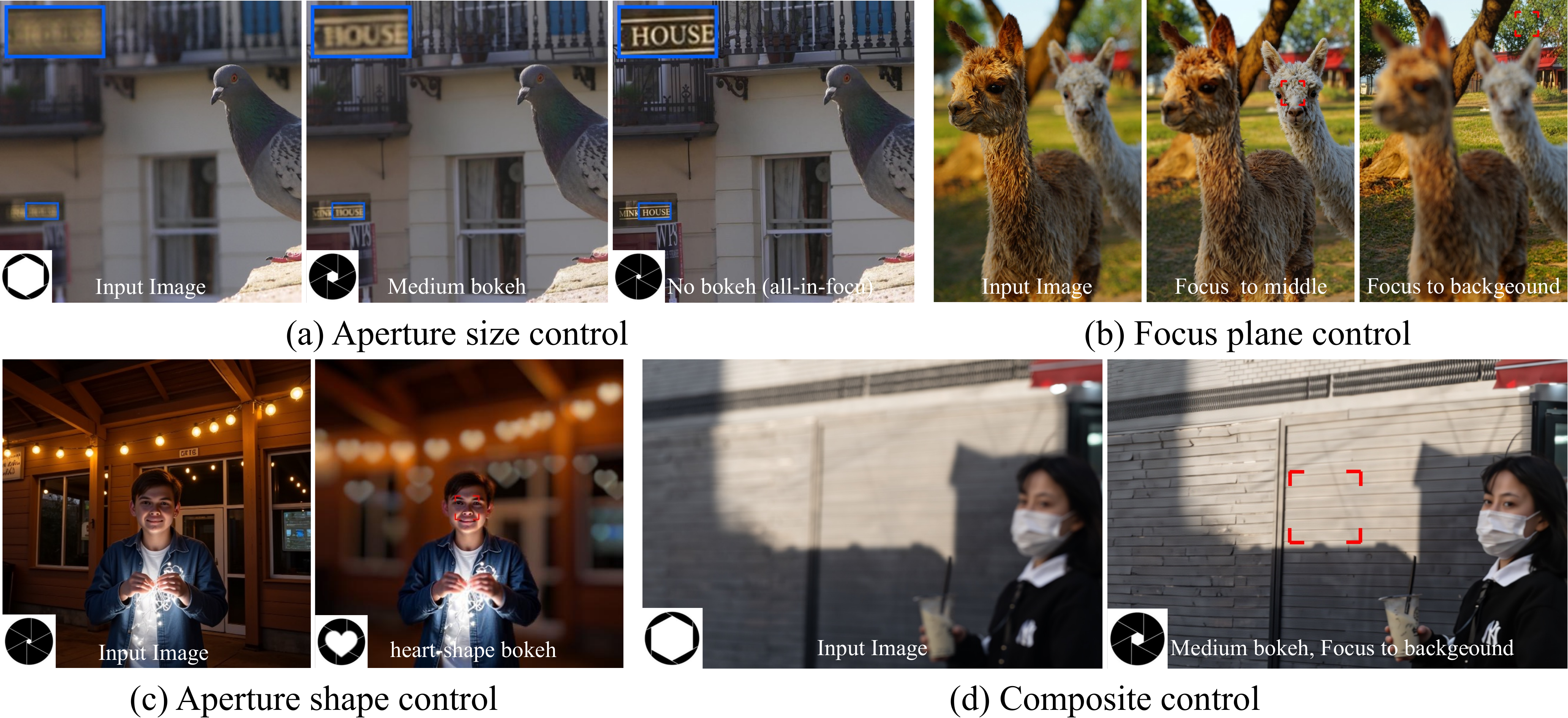}
  \captionsetup{type=figure}
  \caption{\textbf{Generative refocusing with controllable bokeh.}
  Our model turns a single input image into a virtual camera that users can control. This enables a range of adjustments after image capture. (a) demonstrates aperture size control, allowing the user to change the depth of field from strong bokeh to an all-in-focus image. (b) demonstrates focus plane control by shifting the sharp region from the middle subject to the background. (c) highlights aperture-shape control, synthesizing a creative heart-shaped bokeh from point lights in the scene. (d) shows composite control, where both the focus plane and aperture size are adjusted simultaneously to reframe the subject.}
  \label{fig:teaser}
\end{center}

\begin{abstract}
  Depth-of-field control is essential in photography, but achieving perfect focus often requires multiple attempts or specialized equipment. Single-image refocusing is still difficult. It involves recovering sharp content and creating realistic bokeh. Current methods have significant drawbacks. 
  They require all-in-focus inputs, rely on synthetic data from simulators, and have limited control over the aperture. 
  We introduce \textbf{Generative Refocusing}, a two-step process that uses DeblurNet to recover all-in-focus images from diverse inputs and BokehNet to create controllable bokeh. 
  This method combines synthetic and real bokeh images to achieve precise control while preserving authentic optical characteristics.
  Our experiments show we achieve top performance in defocus deblurring, bokeh synthesis, and refocusing benchmarks. 
  Additionally, our Generative Refocusing allows custom aperture shapes.
  Project page: \url{https://generative-refocusing.github.io/}
\end{abstract}

\section{Introduction}
\label{sec:intro}

Controlling depth of field and focus is essential for artistic photography. 
It helps guide the viewer’s eye through selective focus and bokeh effects. 
Achieving perfect focus often requires multiple shots or specialized equipment, which can be challenging for everyday photographers. Single-image refocusing removes these obstacles. It allows for focus and bokeh adjustments \emph{after} the photo is taken. The main challenge is recovering sharp details from blurred areas while creating natural-looking bokeh, while still giving users fine control.

Existing research has addressed various aspects of this problem. Defocus deblurring methods~\cite{DPDD,chen2025quad,IFANet,DRBNet,tsai2022stripformer,wang2022uformer,Restormer} aim to recover sharp images. Diffusion approaches~\cite{li2025real,liang2025swin} show promise for addressing spatially variant blur. Bokeh synthesis methods~\cite{chen2024variable,ebb,BokehMe,Bokehlicious} focus on rendering realistic depth-of-field. Recent efforts include physics-based~\cite{lee2008real,lee2009depth,potmesil1981lens}, neural~\cite{BokehMe,peng2024bokehme++}, and diffusion-based~\cite{Fortes2025BokehDiffusion,yuan2025generative,wang2025diffcamera} approaches. However, most methods handle either deblurring or bokeh synthesis \emph{individually}. End-to-end refocusing often requires specialized capture setups~\cite{levoy2023light,ng2005light,DC2}.

Despite these advances, three main limitations remain (\cref{tab:method_comparison}). First, most methods rely on all-in-focus inputs and accurate depth maps. 
This restricts their use to images that already have defocus blur. Second, while synthetic data~\cite{BokehMe,BokehDiff, wang2025diffcamera} can be used for training, its realism is limited by the fidelity of the simulators. This approach often fails to capture detailed visual appearances. On the other hand, real datasets~\cite{Bokehlicious, tedla2025learning} suffer from a critical limitation: they are restricted to isolated one-dimensional variations (i.e., solely aperture sweeps or focus plane sweeps). Consequently, they fail to provide the joint distribution of focus and aperture changes required to train a model for flexible, dual-parameter refocusing. 
Third, current methods typically support only aperture size, not its shape. 
This restriction limits the possibilities for creative bokeh effects.

\definecolor{darkgreen}{rgb}{0.0, 0.5, 0.0}
\newcommand{\marksize}{\LARGE}
\newcommand{\cmark}{\textcolor{darkgreen}{\large\ding{51}}}%
\newcommand{\xmark}{\textcolor{red}{\large\ding{55}}}%
\newcommand{\maybe}{\ensuremath{\mathlarger{\triangle}}}

\begin{table*}[t]
  \centering
  \setlength{\tabcolsep}{6pt}
  \renewcommand{\arraystretch}{1.12}
\caption{\textbf{Comparison across methods.} \cmark~= supported; \xmark~= not supported; \maybe~= supported with caveat. Flexible input indicates compatibility with both all-in-focus (AIF) and defocused images. BokehMe and Bokehdiff require AIF inputs, while Learn2refocus strictly uses defocused inputs.
$^\star$For BokehMe, only the \textit{classical renderer} variant supports aperture-shape control. $^\dagger$Bokehlicious typically assumes an all-in-focus input and requires fine-tuning for different input types. $^\ddagger$Learn2Refocus generates focal stacks at discrete depth planes rather than supporting point-based focus selection.}
\label{tab:method_comparison}
\resizebox{\textwidth}{!}{
\begin{tabular}{l c c c c c c c c c}
  \toprule
  \multirow{2}{*}[-1.5ex]{\textbf{Method}} &
    \multirow{2}{*}[-1.5ex]{\textbf{\shortstack{Flexible\\[1pt]input?}}} &
    \multirow{2}{*}[-1.5ex]{\textbf{\shortstack{Synthetic\\training data}}} &
    \multicolumn{3}{c}{\textbf{Real training data}} &
    \multicolumn{4}{c}{\textbf{Supported user control types}} \\
  \cmidrule(lr){4-6} \cmidrule(lr){7-10}
   & & & 
     \textbf{\shortstack{Focus-\\[1pt]sweep}} & \textbf{\shortstack{Aperture-\\[-1pt]sweep}} & \textbf{\shortstack{Single\\[-3pt]images}} & 
     \textbf{\shortstack{Aperture\\[-3pt]size}} & \textbf{\shortstack{Aperture\\[-3pt]shape}} & \textbf{\shortstack{Focus\\[-1pt]point}} & \textbf{\shortstack{Defocus\\map}} \\
  \midrule
  BokehMe \cite{BokehMe}                     & \xmark & \cmark & \xmark & \xmark & \xmark & \cmark & \maybe$^\star$ & \cmark & \cmark \\
  Bokehlicious$^\dagger$ \cite{Bokehlicious} & \cmark & \xmark & \xmark & \cmark & \xmark & \cmark & \xmark & \xmark & \xmark \\
  BokehDiff \cite{BokehDiff}                 & \xmark & \cmark & \xmark & \xmark & \xmark & \cmark & \xmark         & \cmark & \cmark \\
  DiffCamera \cite{wang2025diffcamera}       & \cmark & \cmark & \xmark & \xmark & \xmark & \cmark & \xmark         & \cmark & \xmark \\
  Learn2Refocus \cite{tedla2025learning}     & \xmark & \xmark & \cmark & \xmark & \xmark & \xmark & \xmark         & \maybe$^\ddagger$ & \xmark \\

  GenRefocus (Ours)                                       & \cmark & \cmark & \cmark & \cmark & \cmark & \cmark & \cmark         & \cmark & \cmark \\
  \bottomrule
\end{tabular}
}
\end{table*}

To tackle these issues, we present \textbf{Generative Refocusing (GenRefocus)}, a flexible single-image refocusing system built on a two-stage design. Our DeblurNet module produces a sharp, in-focus image from a variety of inputs. It uses a diffusion model guided by initial deblurring predictions. 
Our BokehNet generates fully customizable bokeh, accounting for user-defined focus planes, bokeh intensity, and aperture shapes (\cref{fig:teaser}). 
A significant advancement of our method is the proposed training scheme. 
While we utilize synthetic data to maintain geometric consistency, our key contribution lies in handling real-world data. By leveraging auxiliary information and dedicated processing, we complete the missing input or control signals (i.e., defocus maps) in existing real-world datasets. This strategy enables our model to effectively learn authentic lens characteristics that simulators fail to capture. Our approach achieves top performance in defocus deblurring, bokeh generation, and refocusing, while maintaining the natural consistency of scenes.


Our main contributions are:
\begin{itemize}
\item A flexible refocusing pipeline accepts images in any focus state. 
It provides users with control over the focus plane, bokeh intensity, and aperture shape via two-stage decomposition. 
\item A novel training scheme that overcomes the limitations of existing real-world datasets by completing absent input or control signals. This strategy successfully unites the geometric consistency of synthetic data with the authentic aberrations of real lenses.
\item The system performs well across defocus deblurring, bokeh synthesis, and refocusing. This has been validated on existing benchmarks (DPDD~\cite{DPDD}, RealDOF~\cite{IFANet}) and our new light-field datasets (LF-Bokeh, LF-Refocus). It also has applications in creating aperture shapes.
\end{itemize}

\section{Related Work}

\begin{figure*}[t]
  \centering
  \includegraphics[width=\textwidth]{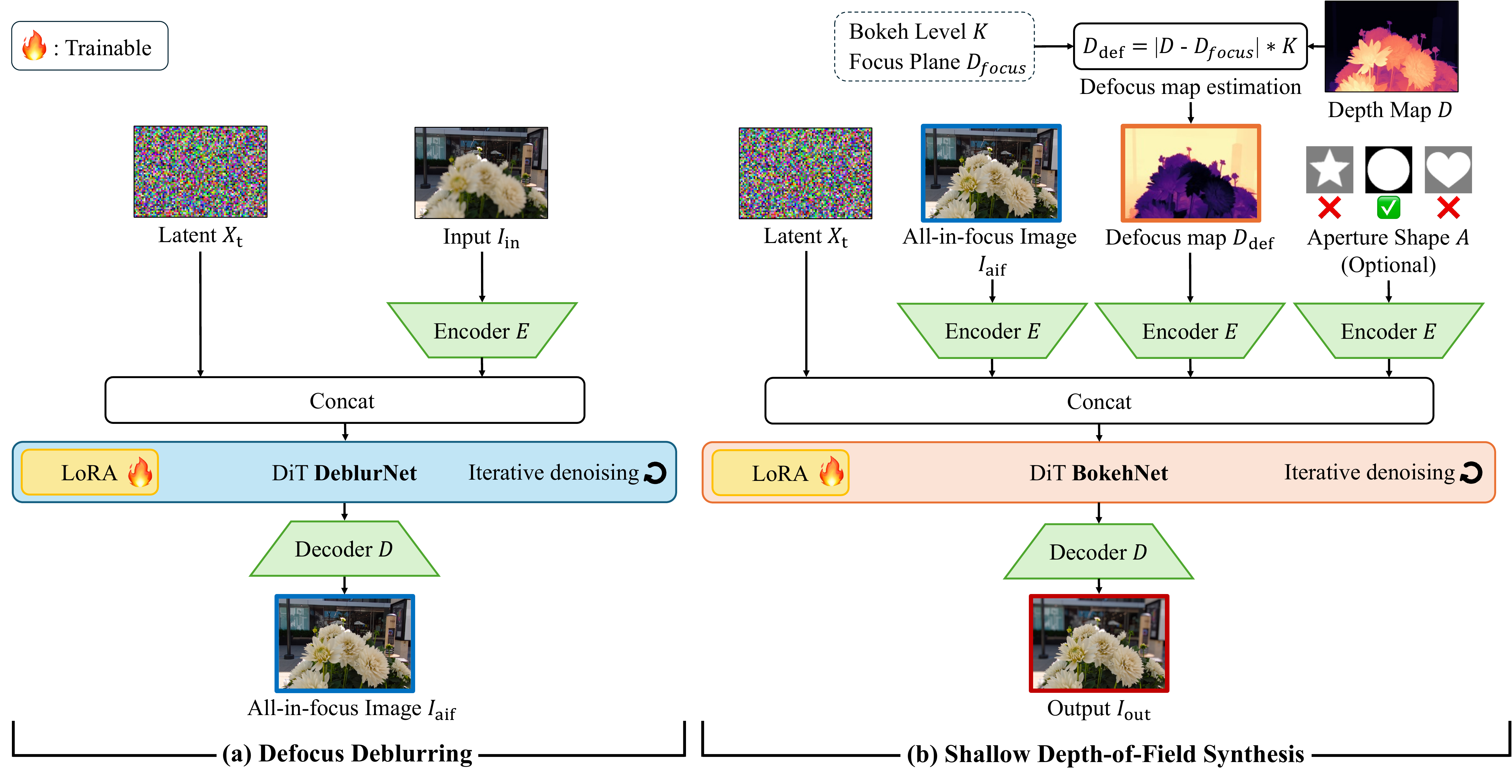}
\caption{\textbf{Pipeline Overview.} Our method decomposes single-image refocusing into two stages:
  \textbf{(a) Defocus Deblurring} and \textbf{(b) Shallow Depth-of-Field Synthesis}.
  \textbf{(a)} The noisy latent $X_t$ and the encoded blurry input $I_{\text{in}}$ are concatenated into a unified token sequence $\mathbf{S}_t$. DeblurNet iteratively denoises this sequence to reconstruct a high-quality all-in-focus image $I_{\text{aif}}$.
  \textbf{(b)} Following the same unified-sequence design, BokehNet conditions on $I_{\text{aif}}$ and the defocus map $D_{\text{def}}$ to synthesize the refocused output $I_{\text{out}}$, with an optional aperture-shape condition $A$.
  The VAE encoder $\mathcal{E}$ and decoder $\mathcal{D}$ map images to latent representations for the DiT backbone.
  The defocus map $D_{\text{def}}$ is computed from the estimated depth map $D$~\cite{Bochkovskiy2025DepthPro} together with the user-specified focus plane $D_\text{focus}$ and bokeh level $K$.
}
\label{fig:pipeline}
\end{figure*}

\subsubsection{Diffusion Models for Image Restoration.}
Diffusion models~\cite{ho2020denoising,song2020denoising,yeh2024diffir2vr,tsai2025lightsout} advance image restoration via generative priors, extending to zero-shot restoration and artifact removal.
Methods evolved from pixel-space~\cite{saharia2022image} to efficient latent-space approaches~\cite{rombach2022high,xia2023diffir}, enabling faster inference.
ResShift~\cite{yue2023resshift} cuts steps from 1000 to 4–15 via residual shifting.
Deblurring has progressed from merging diffusion priors with regression tasks, such as matting~\cite{wang2024matting} or hierarchical restoration~\cite{chen2023hierarchical}, to learning spatially varying kernels in latent space~\cite{kong2025deblurdiff}.
Cascaded pipelines~\cite{ho2022cascaded} refine at multiple resolutions, while principled frameworks~\cite{luo2023image,luo2025visual} treat degradations as stochastic processes.
Training-free methods~\cite{zhu2023denoising,kawar2022denoising} enable plug-and-play restoration.
\emph{Unlike} these general methods, we target \emph{spatially-varying defocus blur} with a \emph{two-stage pipeline} explicitly separating deblurring from bokeh synthesis for controllable refocusing.

\subsubsection{Defocus Deblurring.}
Defocus deblurring recovers sharp images from spatially varying blur caused by limited depth of field.
Early deconvolution~\cite{bando2007towards,zhuo2011defocus} introduced artifacts at depth transitions.
Dual-pixel sensors~\cite{DPDD}, quad-pixel data~\cite{chen2025quad}, and disparity-aware techniques~\cite{IFANet,yang2023k3dn,wang2021bridging} improved blur–depth modeling and the recovery of all-in-focus images from focal stacks.
Architectures shifted from CNNs~\cite{chen2022simple} to transformers~\cite{tsai2022stripformer,wang2022uformer}, with gains from implicit representations~\cite{INIKNet,fan2025spectromotion} and multi-scale attention~\cite{zhang2024unified}.
Recent work includes vision-language fusion~\cite{yang2024ldp} for semantic estimation and a shift from supervised~\cite{ren2025reblurring} to diffusion-based learning~\cite{liang2025swin,li2025real}, including unpaired data.
\emph{While} these excel as \emph{standalone solutions}, we position deblurring as the \emph{first stage} of refocusing, using FLUX's~\cite{ShakkerLabs2024FluxControlNetUnionPro} generative prior to enable controllable bokeh synthesis from any input.

\subsubsection{Bokeh Rendering.}
Bokeh rendering progressed from physically based scattering~\cite{potmesil1981lens,lee2008real,lee2009depth} and differentiable rendering~\cite{sheng2024dr} to neural methods combining ray tracing with learned enhancements~\cite{BokehMe,peng2024bokehme++}.
Learning-based approaches evolved from fixed apertures~\cite{ebb} to variable f-stops~\cite{Bokehlicious,chen2024variable}, extending to video~\cite{yang2025any} and 3D scenes~\cite{shen2025dof,wang2025dof, huang2025bokehflow}. Diffusion models enabled bokeh control~\cite{Fortes2025BokehDiffusion,yuan2025generative}, mainly for text-to-image synthesis.
These methods require \emph{all-in-focus inputs} and \emph{accurate depth}. 
We relax this constraint via \emph{flexible input handling}, which accepts defocused or fully in-focus images.

\subsubsection{Single-Image Refocusing.}
Post-capture refocusing evolved from deconvolution~\cite{bando2007towards,zhang2011single} and GANs~\cite{sakurikar2018refocusgan} to recent diffusion models~\cite{qin2025camedit}, yet significant limitations remain (see \cref{tab:method_comparison}). Constrained by conditioning on explicit focus coordinates, DiffCamera~\cite{wang2025diffcamera} requires fixed-resolution inputs ($512 \times 512$) and suffers performance drops on arbitrary aspect ratios. Furthermore, it inherits simulator artifacts from its purely synthetic training. Learn2Refocus~\cite{tedla2025learning} trains on real data but lacks aperture-size control and generates a focal stack at predefined depth planes that may not align with the user's target focus. Additionally, 3D representations~\cite{shen2025dof,wang2025dof,lee2024deblurring,liu2023robust} offer controllable DoF but require multi-view capture.
To address these gaps, we \emph{combine three capabilities}: (1) \emph{Hybrid training scheme} mixing synthetic and real data for authentic optics, (2) \emph{single-image flexibility} accepting any input without preprocessing, and (3) \emph{comprehensive control} over focus, intensity, and aperture shape.

\subsubsection{Camera-Conditioned Diffusion.}
Conditioning diffusion models on camera parameters progressed from extrinsic control for view synthesis~\cite{hollein2024viewdiff,cheng2024learning} and video generation~\cite{he2024cameractrl,wang2024motionctrl}, to intrinsic control of focal length, aperture, and exposure~\cite{fang2024camera,chen2023learning}. Direct conditioning often yields inconsistent content. Recent work uses temporal modeling~\cite{yuan2025generative}, explicit calibration~\cite{bernal2025precisecam,deng2025boost}, or transformer-based pose handling~\cite{bahmani2024vd3d,saxena2023zero}.
These focus on \emph{generation}; we operate in \emph{editing}, manipulating images with scene consistency. By \emph{decoupling} deblurring and bokeh stages, we achieve efficient single-image manipulation without multi-frame overhead.


\section{Method}
Single-image refocusing couples two operations: recovering sharp content in out-of-focus regions and applying controllable bokeh to originally sharp areas.
Blur magnitude is dictated by a defocus map parameterized by a user-specified focus point and the scene depth.
When the input is blurry, monocular depth estimation is brittle, undermining precise control over the defocus map and often yielding misfocus or artifacts.
We propose Generative Refocusing (GenRefocus), which decomposes the task into defocus deblurring and bokeh synthesis, and comprises two models, DeblurNet and BokehNet, that jointly enable precise and controllable refocusing (see Fig.~\ref{fig:pipeline} for an overview of the pipeline); 
Such a flexible pipeline allows us to exert greater control over input conditions (see Tab.~\ref{tab:method_comparison}).

\begin{figure*}[t]
  \centering
  \includegraphics[width=\textwidth]{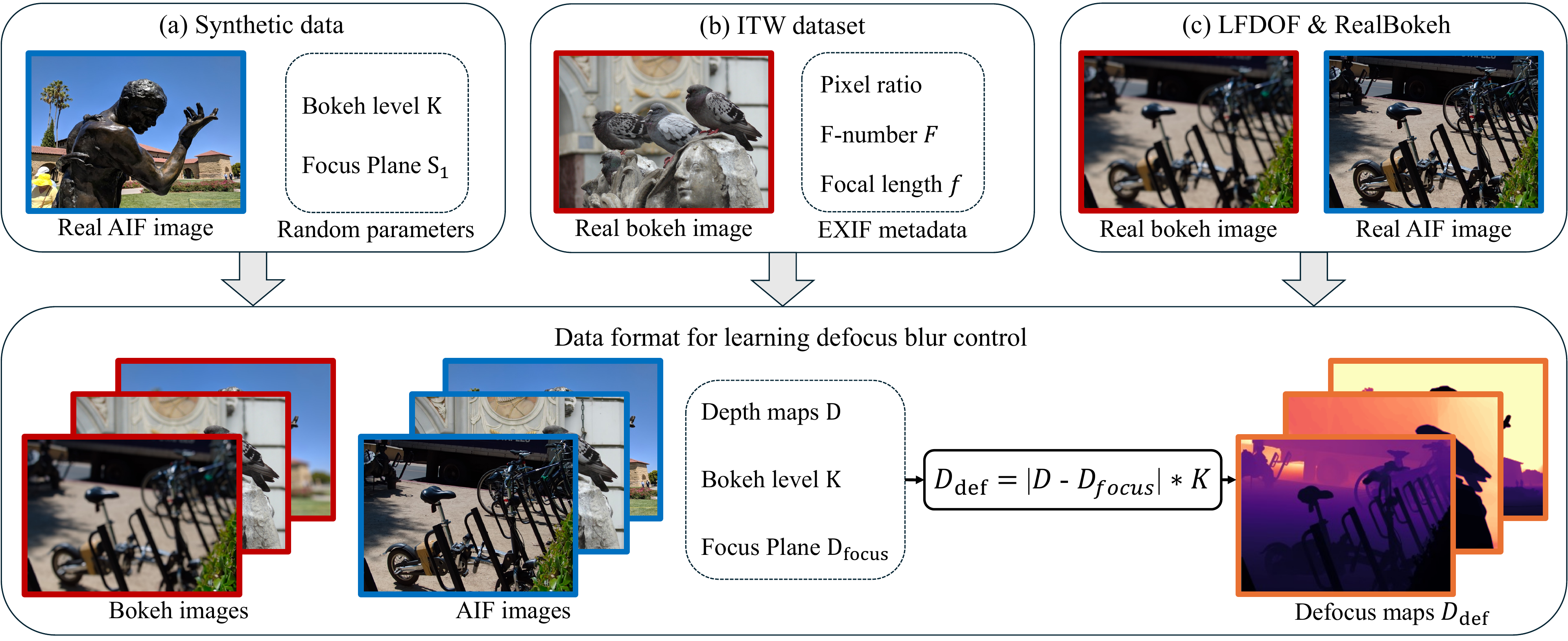}
  \caption{
    \textbf{Training data generation.}
    Each training sample consists of five components:
    (i) a bokeh image,
    (ii) an all-in-focus (AIF) image,
    (iii) a depth map $D$,
    (iv) a bokeh level $K$, and
    (v) a focus plane $D_\text{focus}$.
    We construct these samples via three routes:
    (a) Synthetic data.
    Given real AIF images and depth maps $D$, we compute a defocus map $D_{\text{def}}$ parameterized by randomly selected bokeh level $K$ and focus plane $D_\text{focus}$, and feed it into a bokeh renderer~\cite{BokehMe} to synthesize corresponding bokeh images.
    (b) ITW dataset~\cite{Fortes2025BokehDiffusion}.
    Given real bokeh images, DeblurNet recovers an AIF image.
    We then estimate depth and extract a foreground mask~\cite{Zheng2024BiRefNet} to define the estimated focus plane $D_\text{focus}$.
    The bokeh level $K$ is computed from the EXIF metadata and the estimated $D_\text{focus}$ following the formulation in~\cite{Fortes2025BokehDiffusion}.
    (c) LFDOF~\cite{AIFNet} and RealBokeh.
    For real pairs, we obtain $D_\text{focus}$ as in (b), and follow Eq.~(2) to estimate the bokeh level $K$.
    \label{fig:data_generation}
  }
\end{figure*}

\subsection{Stage 1: Defocus Deblurring}
Given a blurry input image $I_{\text{in}}$, we follow the conditioning design of~\cite{tan2025ominicontrol,Tan2025OminiControl2}.
At each denoising step $t$, DeblurNet processes a single unified token sequence
\begin{equation} \small
\mathbf{S}_t = \bigl[\, X_t \,;\, \mathcal{E}(I_{\text{in}}) \,\bigr],
\end{equation}
where $X_t$ denotes the noisy image latent tokens at timestep $t$, $\mathcal{E}$ is a pretrained VAE encoder that maps the input image to latent condition tokens, and $[\,\cdot\,;\,\cdot\,]$ denotes token concatenation.
After iterative denoising, we decode the image latent part with the VAE decoder $\mathcal{D}$ to obtain the all-in-focus output $I_{\text{aif}}$ (Fig.~\ref{fig:pipeline}\,(a)).
We train DeblurNet in a supervised manner using real paired data~\cite{DPDD,Bokehlicious}.

\subsection{Stage 2: Shallow Depth-of-Field Synthesis}
\label{sec:bokehnet}
Given an all-in-focus image $I_{\text{aif}}$ and a user-specified focus plane $D_\text{focus}$ with bokeh level $K$, we define the defocus map $D_{\text{def}}$ as
\begin{equation}\small
D_{\text{def}} = K \cdot \left| D - D_\text{focus} \right|,
\label{eq:defocus_map}
\end{equation}
where $D$ is the monocular depth map estimated from $I_{\text{aif}}$ using an off-the-shelf depth estimator~\cite{Bochkovskiy2025DepthPro}.
Following the unified-token conditioning design in Stage~1, BokehNet concatenates the condition tokens of $I_{\text{aif}}$ and $D_{\text{def}}$ with the noisy latent tokens at each denoising step.
BokehNet then iteratively denoises $X_t$ to synthesize the refocused output $I_{\text{out}}$, as illustrated in Fig.~\ref{fig:pipeline}\,(b).

\subsubsection{Training data and supervision.}
Training BokehNet requires $(I_{\text{aif}}, I_{\text{out}}, D_{\text{def}})$.
Because accurately paired real supervision is scarce, we construct supervision via three complementary sources (Fig.~\ref{fig:data_generation}): (a) synthetic data, (b) ITW dataset\cite{Fortes2025BokehDiffusion}, and (c) LFDOF\cite{AIFNet} and RealBokeh\cite{Bokehlicious}.

\paragraph{(a) Synthetic data.}
We pretrain with synthetic data: starting from real all-in-focus images and their estimated depth map $D$, we randomly sample a focus plane $D_\text{focus}$ and a target bokeh level $K$ (Fig.~\ref{fig:data_generation}\,(a)).
We then construct $D_{\text{def}}$ using Eq.~\ref{eq:defocus_map} and use a simulator~\cite{BokehMe} to render the corresponding target bokeh image consistent with $D_{\text{def}}$.
This synthetic pretraining helps the network modulate the circle of confusion according to $D_{\text{def}}$; however, it is constrained by renderer bias and may introduce unrealistic artifacts.

\paragraph{(b) ITW dataset.}
To capture real optics, we leverage ITW dataset\cite{Fortes2025BokehDiffusion} contains real bokeh images (see Fig.~\ref{fig:data_generation}\,(b)).
For each real image, we use DeblurNet to produce the AIF input and compute an \emph{approximate} bokeh level $\!K$ following \cite{Fortes2025BokehDiffusion}:
\begin{equation} \small
\label{eq:approxK}
K \;\approx\; \frac{f^{2}\,D_\text{focus}}{\,2\,F\,(D_\text{focus}-f)\,}\times \text{pixel\_ratio},
\end{equation}
where $f$ is the focal length and $F$ is the aperture $f$-number, both directly obtained from EXIF metadata. 
The pixel\_ratio term accounts for differences in camera sensor size and image resolution.
Although some devices may provide a focus-distance field in EXIF, it is frequently missing or noisy; therefore, we \emph{do not} rely on EXIF for $D_\text{focus}$.
Instead, we compute the focus plane based on an in-focus mask and a monocular depth estimate, following the approach of ~\cite{Fortes2025BokehDiffusion}.
Specifically, we estimate an in-focus mask $M$ using BiRefNet~\cite{Zheng2024BiRefNet} and produce a depth map $D$ using a monocular depth estimator.
The focus plane is then determined as the median depth within the masked in-focus area:
\begin{equation}\small
D_\text{focus} = \mathrm{median}\bigl(D[M]\bigr).
\label{eq:focus_plane_median}
\end{equation}
We then generate $D_{\text{def}}$ using Eq.~\ref{eq:defocus_map}, enabling BokehNet to effectively train on real single images. While these estimated AIF images and defocus maps contain inevitable noise, training on them inherently aligns with our single-image refocusing pipeline, making the model more robust to depth and deblurring artifacts.

\paragraph{(c) LFDOF and RealBokeh.}
These datasets provide pairs but omit EXIF metadata or provide insufficient fields to estimate $K$ (see Fig.~\ref{fig:data_generation}\,(c)). 
To address this, we first compute the focus-plane proxy $D_\text{focus}$. Similar to (b), we employ BiRefNet~\cite{Zheng2024BiRefNet} to obtain an initial in-focus mask $M$. However, due to the increased diversity and complexity of the scenes in these datasets, the initial estimate of $M$ is sometimes unreliable. 
Rather than simply verifying and discarding unreliable cases, we introduce a manual refinement step. Specifically, we re-select a small yet reliable in-focus region to correct $M$, thereby accurately extracting $D_\text{focus}$. 
This strategy preserves challenging samples rather than excluding them, thereby enabling the model to learn from a wider range of real-world scenarios.

With $D_\text{focus}$ established, we adopt a \emph{simulator-in-the-loop calibration} of the bokeh level. Given an AIF image $I_{\mathrm{aif}}$ and estimated depth $D$, we sweep $K$ and choose the value whose rendered result best matches the real bokeh target $I_{\mathrm{real}}$:
\begin{equation}\small
K^{\star} = \operatorname*{argmax}_{\substack{K \in \\ (K_{\min},\, K_{\max})}}
\operatorname{SSIM}\!\big(R(I_{\mathrm{aif}}, D;\, D_\text{focus}, K),\, I_{\mathrm{real}}\big),
\end{equation}
where $R$ denotes our physically guided renderer. The selected $K^{\star}$ is then used as the pseudo-bokeh-level label for training, provided that its corresponding SSIM exceeds a predefined threshold to ensure reliable supervision. Finally, the defocus map $D_{\text{def}}$ is constructed accordingly following Eq.~\ref{eq:defocus_map}.
\subsection{Bokeh-Shape Aware Synthesis}
\label{sec:shapeaware}
State-of-the-art learning-based bokeh synthesis methods deliver compelling results but \emph{do not expose} aperture-shape control.
Physics-based renderers, in principle, support arbitrary shapes, yet public implementations typically omit this functionality.
We therefore explore explicit \emph{shape-aware} control using only a bokeh shape image within a trainable framework.

\begin{table*}[t]
  \centering
  \small
  \setlength{\tabcolsep}{3pt}
  \caption{
  \textbf{Defocus deblurring benchmark on RealDOF~\cite{IFANet} and DPDD~\cite{DPDD} datasets.} We report reference (LPIPS, DISTS) and no-reference (CLIP-IQA, MANIQA, MUSIQ) perceptual metrics. \colorbox{red!25}{Best}, \colorbox{orange!25}{second best}, and \colorbox{yellow!25}{third best} results are highlighted.
}
  \label{tab:realdof_dpdd_fullwidth}
\resizebox{\textwidth}{!}{
  \begin{tabular}{lcccccccccc}
    \toprule
    & \multicolumn{5}{c}{{RealDOF~\cite{IFANet}}} & \multicolumn{5}{c}{{DPDD~\cite{DPDD}}} \\
    \cmidrule(lr){2-6} \cmidrule(lr){7-11}
    {Method} &
    LPIPS $\downarrow$ & DISTS $\downarrow$ & CLIP\text{-}IQA $\uparrow$ & MANIQA $\uparrow$ & MUSIQ $\uparrow$ &
    LPIPS $\downarrow$ & DISTS $\downarrow$ & CLIP\text{-}IQA $\uparrow$ & MANIQA $\uparrow$ & MUSIQ $\uparrow$ \\
    \midrule
    Input                    & 0.5241 & 0.2865 & 0.3562 & 0.2213 & 28.7087 & 0.3485 & 0.1827 & 0.4337 & 0.3325 & 45.5376 \\
    \midrule
    DRBNet~\cite{DRBNet}     & \cellcolor{yellow!25}0.2550 & \cellcolor{yellow!25}0.1312 & 0.3889 & 0.2609 & 38.9014 & 0.1819 & \cellcolor{yellow!25}0.1063 & 0.4142 & 0.3254 & 46.7229 \\
    Restormer~\cite{Restormer} & 0.2863 & 0.1573 & \cellcolor{yellow!25}0.4062 & 0.2642 & \cellcolor{yellow!25}40.8355 & \cellcolor{yellow!25}0.1762 & 0.1204 & \cellcolor{yellow!25}0.4332 & \cellcolor{yellow!25}0.3307 & \cellcolor{orange!25}47.9872 \\
    INIKNet~\cite{INIKNet}   & 0.2851 & 0.1615 & 0.3984 & 0.2580 & 38.9959 & 0.1860 & 0.1235 & 0.4290 & 0.3265 & 47.1961 \\
    Bokehlicious~\cite{Bokehlicious}   & \cellcolor{red!25}0.2080 & \cellcolor{red!25}0.1070 & \cellcolor{orange!25}0.4281 & \cellcolor{orange!25}0.2746 & \cellcolor{orange!25}42.2314 & \cellcolor{orange!25}0.1598 & \cellcolor{orange!25}0.0918 & 0.4322 & 0.3237 & \cellcolor{yellow!25}47.6057 \\
    DiffCamera~\cite{wang2025diffcamera}     & 0.4523 & 0.2140 & 0.3861 & \cellcolor{yellow!25}0.2667 & 37.9951 & 0.4048 & 0.2141 & \cellcolor{orange!25}0.4438 & \cellcolor{orange!25}0.3350 & 47.1452 \\
    GenRefocus (Ours)& \cellcolor{orange!25}0.2408 & \cellcolor{orange!25}0.1126 & \cellcolor{red!25}0.4595 & \cellcolor{red!25}0.2884 & \cellcolor{red!25}43.5222 &
                                 \cellcolor{red!25}0.1440 & \cellcolor{red!25}0.0772 & \cellcolor{red!25}0.4755 & \cellcolor{red!25}0.3452 & \cellcolor{red!25}49.4122 \\
    \bottomrule
  \end{tabular}
  }
\end{table*}

\begin{figure*}[t]
\centering
\includegraphics[width=\linewidth]{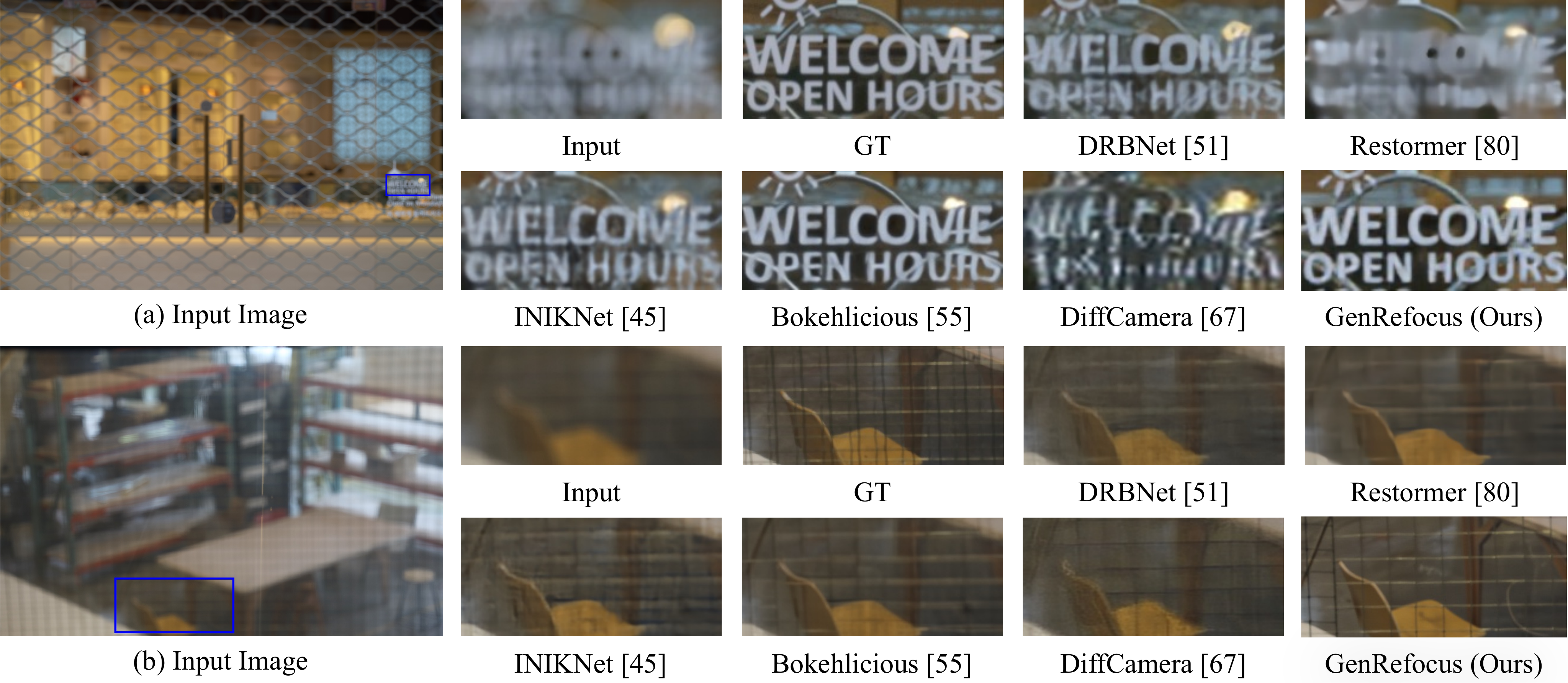}
\caption{
\textbf{Qualitative comparison on defocus deblurring.} Visual results on \textbf{(a)} RealDOF~\cite{IFANet} and \textbf{(b)} DPDD~\cite{DPDD} datasets. 
\textcolor{blue}{Blue} boxes on the left indicate cropped regions shown in detail. In (a), most methods fail to resolve the text, whereas both our method and Bokehlicious~\cite{Bokehlicious} successfully recover the words ``WELCOME'' and ``OPEN HOURS''. However, Bokehlicious produces noticeable distortions in the right-side characters (e.g., ``ME'' and ``RS''), whereas our approach faithfully restores the original text structure. For the challenging example (b) with severe defocus blur, all baseline methods remain highly blurry and fail to recover meaningful details. In contrast, our method synthesizes a relatively clear and visually compelling output.
}
\label{fig:defocus deblur}
\end{figure*}

\subsubsection{Simulator and Data.}
Real photographs exhibiting diverse bokeh shapes are rare, and paired AIF--bokeh examples are even scarcer.
We thus rely on simulation.
A key observation is that when an all-in-focus (AIF) image lacks point-light stimuli, the simulated bokeh carries weak shape cues, making the model reluctant to learn shape-conditioned responses.
To address this, we synthesize a \textbf{PointLight-1K} dataset designed to reveal aperture shape.
We also extend the classical BokehMe renderer to scatter through a given shape:
given an AIF image $I_{\mathrm{aif}}$, depth $D$, focus-plane proxy $D_\text{focus}$, bokeh level $K$, and a shape kernel $s$ (binary/raster PSF), the simulator \emph{R} renders
\begin{equation}\small
\label{eq:shape-sim}
I_{\mathrm{syn}} \;=\; \mathcal{R}\!\big(I_{\mathrm{aif}}, D;\, D_\text{focus}, K, s\big),
\end{equation}
yielding paired supervision for shape-aware training.

\subsubsection{Shape-conditioned fine-tuning.}
To incorporate the aperture shape $A$, we append its tokens directly to the unified sequence. To ensure that shape edits do not regress the learned bokeh synthesis, we freeze all original LoRA weights and introduce a new, trainable LoRA module. Only this dedicated LoRA is fine-tuned to handle the shape conditioning.

\subsection{Pre-deblur Module}
We also study a variant that augments DeblurNet with a conservative pre-deblur prior.
We train this variant using the same data as the base DeblurNet.
Specifically, it condition on $I_{\text{in}}$ and $I_{\text{pd}}$, where $I_{\text{pd}}$ is generated by an off-the-shelf defocus deblurring model.
In the implementation, we use a non-generative defocus model~\cite{DRBNet}, whose outputs are typically content-faithful but overly smooth.
Qualitative results for this variant are shown in the supplementary material.


\subsection{Implementation Details}
While our backbone~\cite{ShakkerLabs2024FluxControlNetUnionPro} intrinsically supports arbitrary resolutions and aspect ratios without resizing, and our conditioning setup is completely coordinate-independent, ultra-high-resolution inputs (e.g., 4K) still incur prohibitive VRAM consumption and quality degradation. We implement a tiling strategy inspired by~\cite{multidiffusion} during inference. Specifically, we process the image as overlapping patches and blend the tile-wise denoised results. This strategy enables us to process images exactly at their original resolutions, fully preserving image fidelity.

\section{Experiments}
\label{sec:Experiments}

\subsection{Setup}
\label{sec:setup}

\subsubsection{Backbone and training details.}
Our backbone is FLUX-1-dev \cite{ShakkerLabs2024FluxControlNetUnionPro}, fine-tuned via LoRA \cite{hu2022lora} with a conditioning scheme following \cite{tan2025ominicontrol,Tan2025OminiControl2}.
DeblurNet employs LoRA rank $r{=}128$, while BokehNet uses $r{=}64$.
Both models are trained with a per-GPU batch size of $1$ and gradient accumulation over $8$ steps on $4{\times}$ RTX~A6000 GPUs.
DeblurNet is trained for $60\mathrm{K}$ steps.
BokehNet is trained in two stages: (i) $40\mathrm{K}$ steps on synthetic data, and (ii) $60\mathrm{K}$ steps on real data.

\subsubsection{Runtime and computational cost.}
During inference, each stage employs 28 denoising steps. 
Without the tiling strategy, our pipeline consumes approximately 40 GB of VRAM and takes approximately 50 seconds per image on a single NVIDIA RTX A6000 GPU.

\subsubsection{Datasets.}
We train \textit{DeblurNet} on 3.5K pairs sourced from DPDD~\cite{DPDD} and a subset of RealBokeh~\cite{Bokehlicious}. For BokehNet, to simultaneously maintain realism and controllability, we utilize a hybrid dataset comprising $\sim$70K synthetic pairs derived from~\cite{yuan2025generative, ebb}, combined with approximately 26K real examples sourced from ITW dataset~\cite{Fortes2025BokehDiffusion}, RealBokeh~\cite{Bokehlicious}, and LFDOF~\cite{AIFNet}.

\begin{table}[t]
\centering
\footnotesize
\caption{
\textbf{Bokeh synthesis benchmark.} 
Evaluation on LF-Bokeh. 
Following~\cite{BokehDiff, BokehMe}, we use per-image binary search for optimal $K$. 
$^\dagger$Since Bokehlicious~\cite{Bokehlicious} uses f-stop rather than $K$ as its control input, we perform a binary search over f-stop values. Due to this fundamentally different control mechanism, we exclude it from the LVCorr evaluation.
$^\ddagger$Note that Bokeh diffusion~\cite{Fortes2025BokehDiffusion} targets the text-to-image task. To adapt it for image editing, we use textual inversion~\cite{song2020denoising}.
}
\label{tab:bokeh}
\setlength{\tabcolsep}{3pt}
\begin{tabular}{lcccc}
\toprule
& \multicolumn{3}{c}{Fidelity} & \multicolumn{1}{c}{Controllability} \\
\cmidrule(lr){2-4} \cmidrule(l){5-5}
Method & LPIPS~$\downarrow$ & DISTS~$\downarrow$ & CLIP-I~$\uparrow$ & LVCorr~$\uparrow$ \\
\midrule
BokehMe \cite{BokehMe}                  & \cellcolor{orange!25}0.1228 & \cellcolor{orange!25}0.0744 & \cellcolor{orange!25}0.9511 & \cellcolor{red!25}0.9940 \\
Bokehlicious$^\dagger$ \cite{Bokehlicious} & 0.1799 & 0.1062 & 0.9304 & - \\
BokehDiff \cite{BokehDiff}                & 0.1708 & 0.0933 & 0.9192 & \cellcolor{yellow!25} 0.8976 \\
Bokeh Diffusion$^\ddagger$  \cite{Fortes2025BokehDiffusion}                & 0.4999 & 0.2529 & 0.7569 & 0.8954 \\
DiffCamera \cite{wang2025diffcamera}                & \cellcolor{yellow!25}0.1429 & \cellcolor{yellow!25}0.0780 & \cellcolor{yellow!25}0.9441 & 0.5632 \\
GenRefocus (Ours) & \cellcolor{red!25}0.0833 & \cellcolor{red!25}0.0487 & \cellcolor{red!25}0.9713 & \cellcolor{orange!25}0.9368 \\
\bottomrule
\end{tabular}
\end{table}

\begin{figure*}[t]
\centering
\small
\setlength{\tabcolsep}{3pt}
\resizebox{\textwidth}{!}{
\begin{tabular}{ccccccc}
\includegraphics[height=0.38\textwidth]{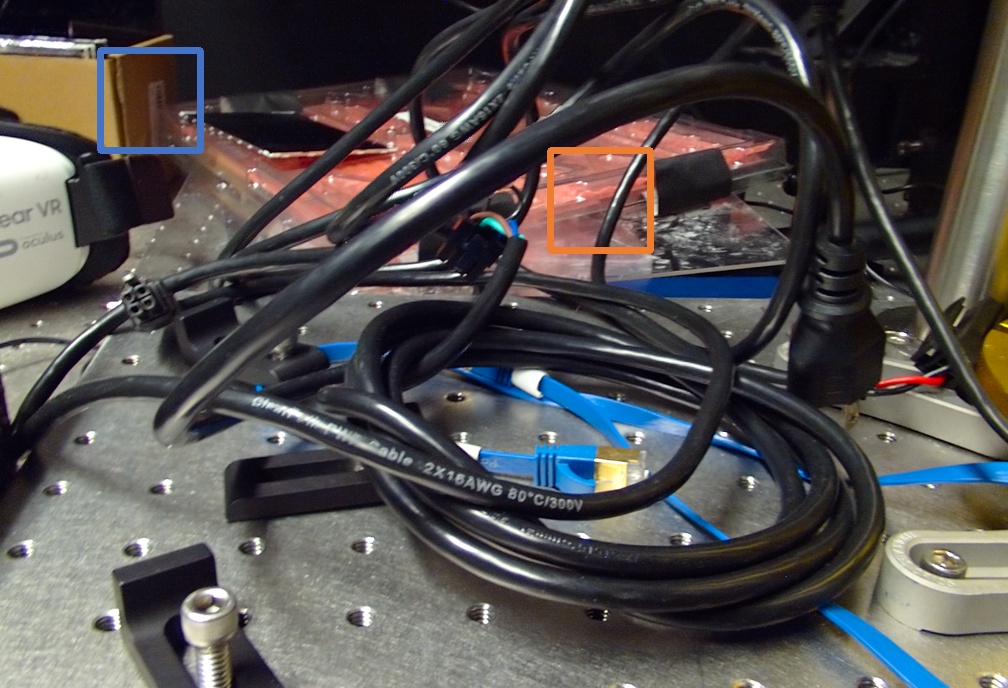} & 
\includegraphics[height=0.38\textwidth]{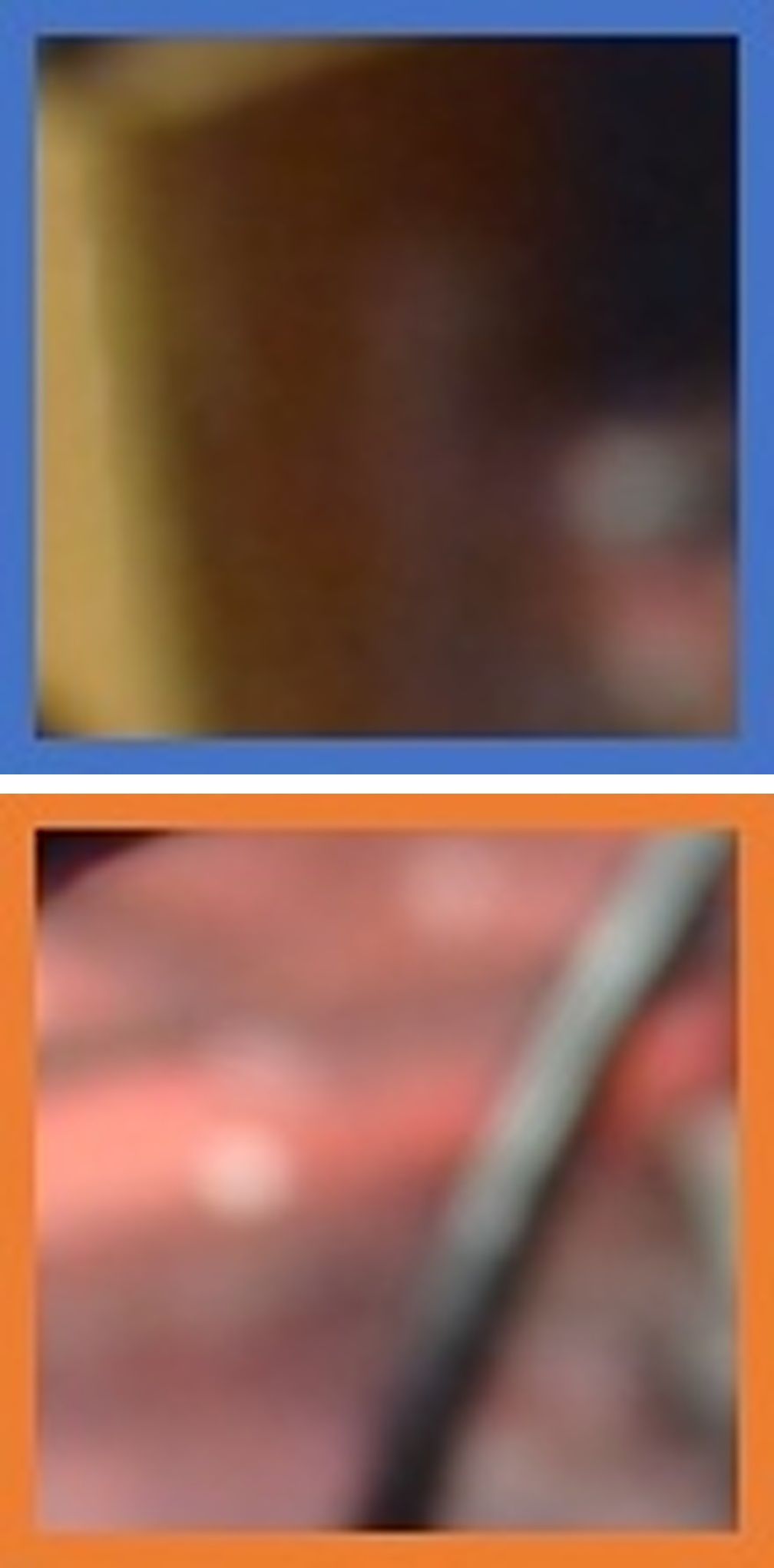} & 
\includegraphics[height=0.38\textwidth]{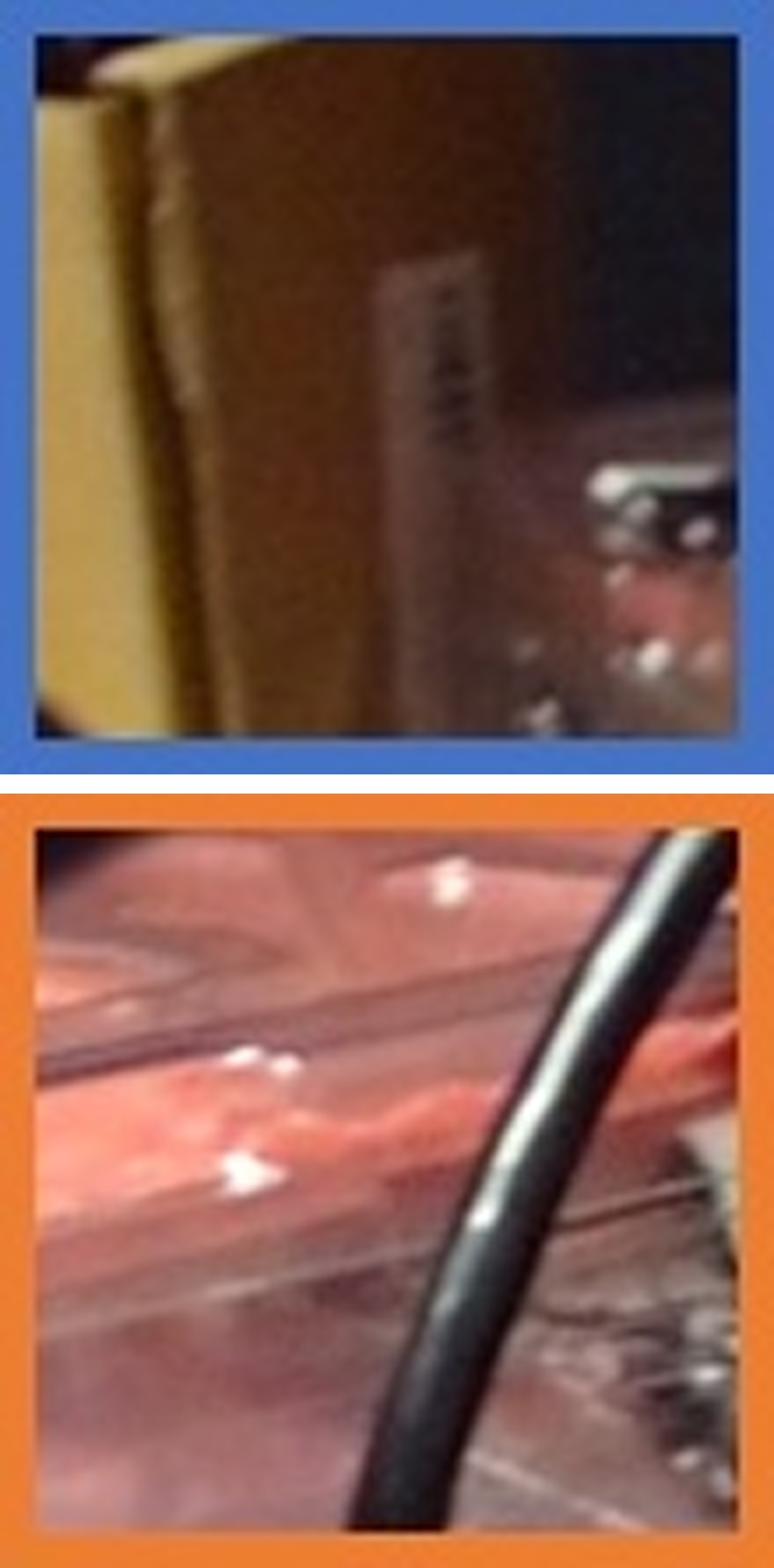} & 
\includegraphics[height=0.38\textwidth]{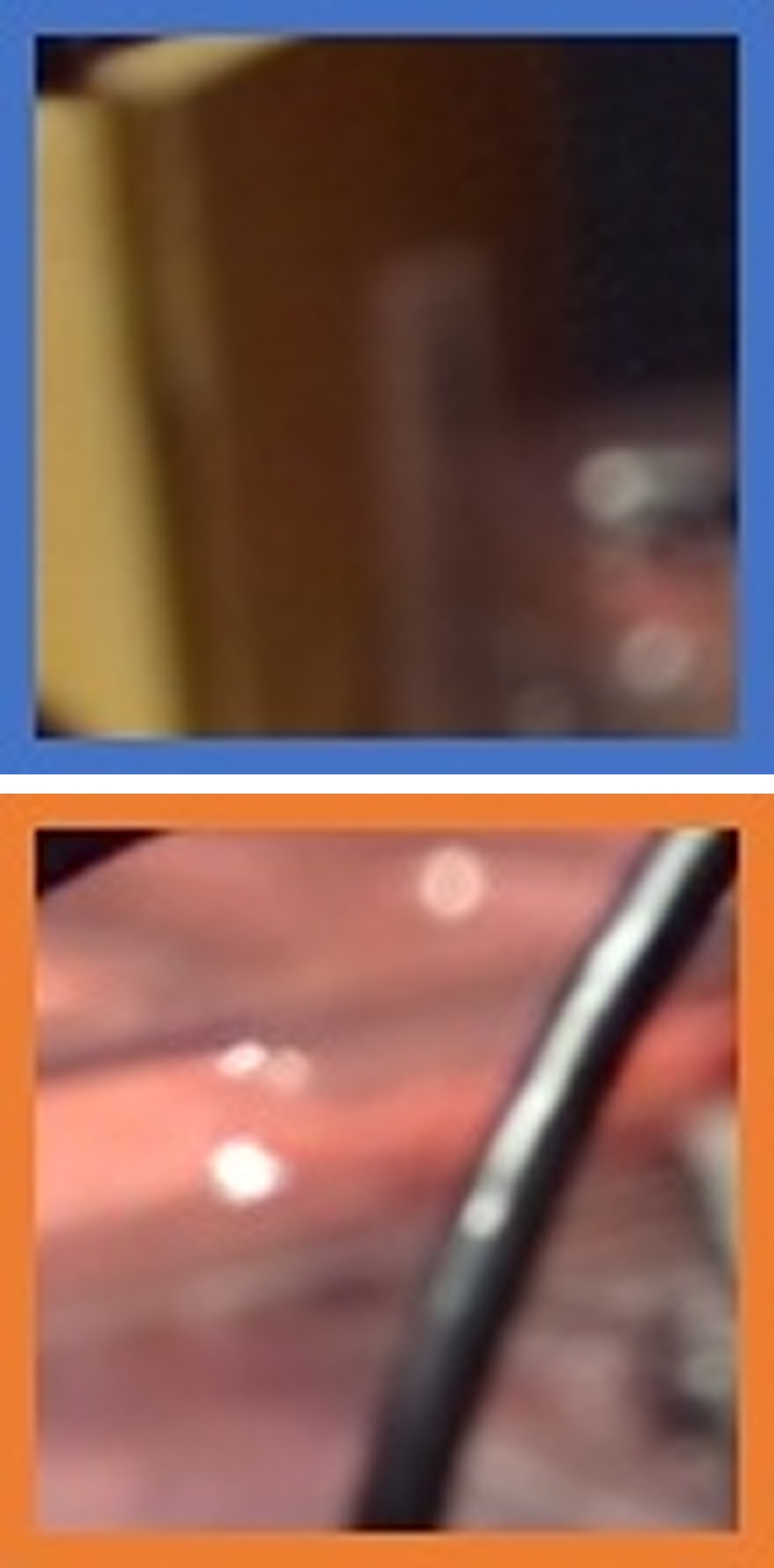} & 
\includegraphics[height=0.38\textwidth]{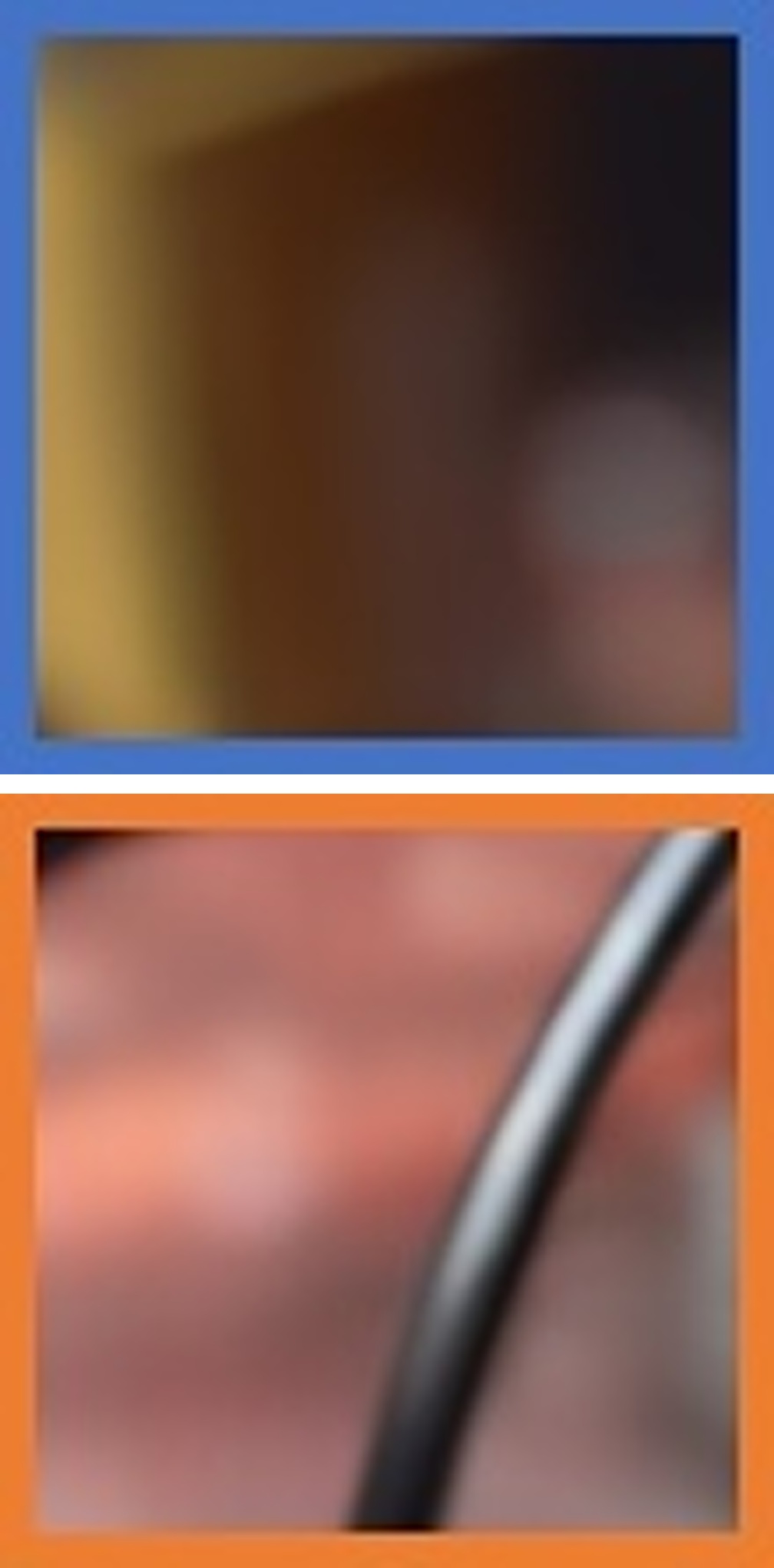} & 
\includegraphics[height=0.38\textwidth]{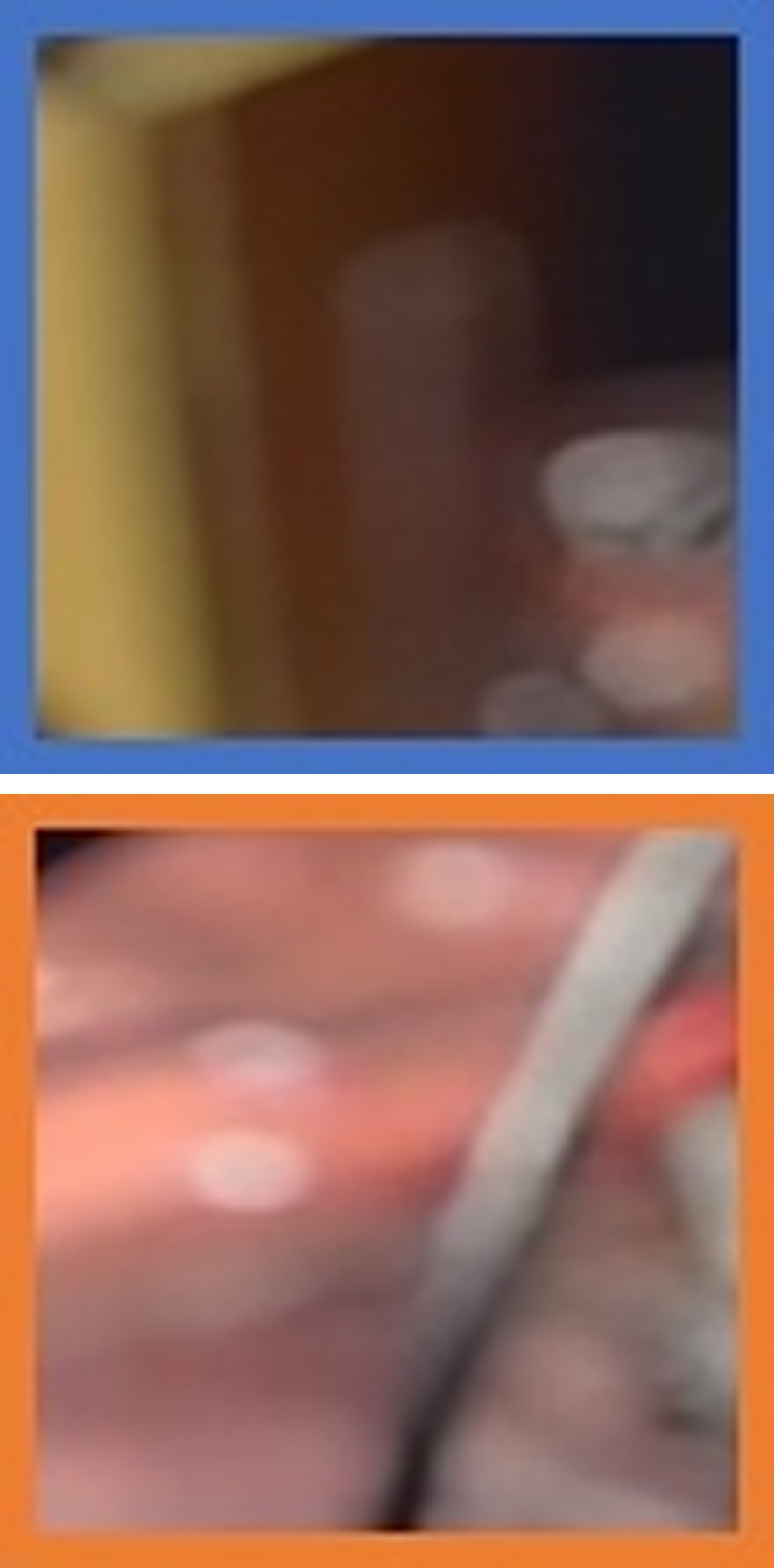} & 
\includegraphics[height=0.38\textwidth]{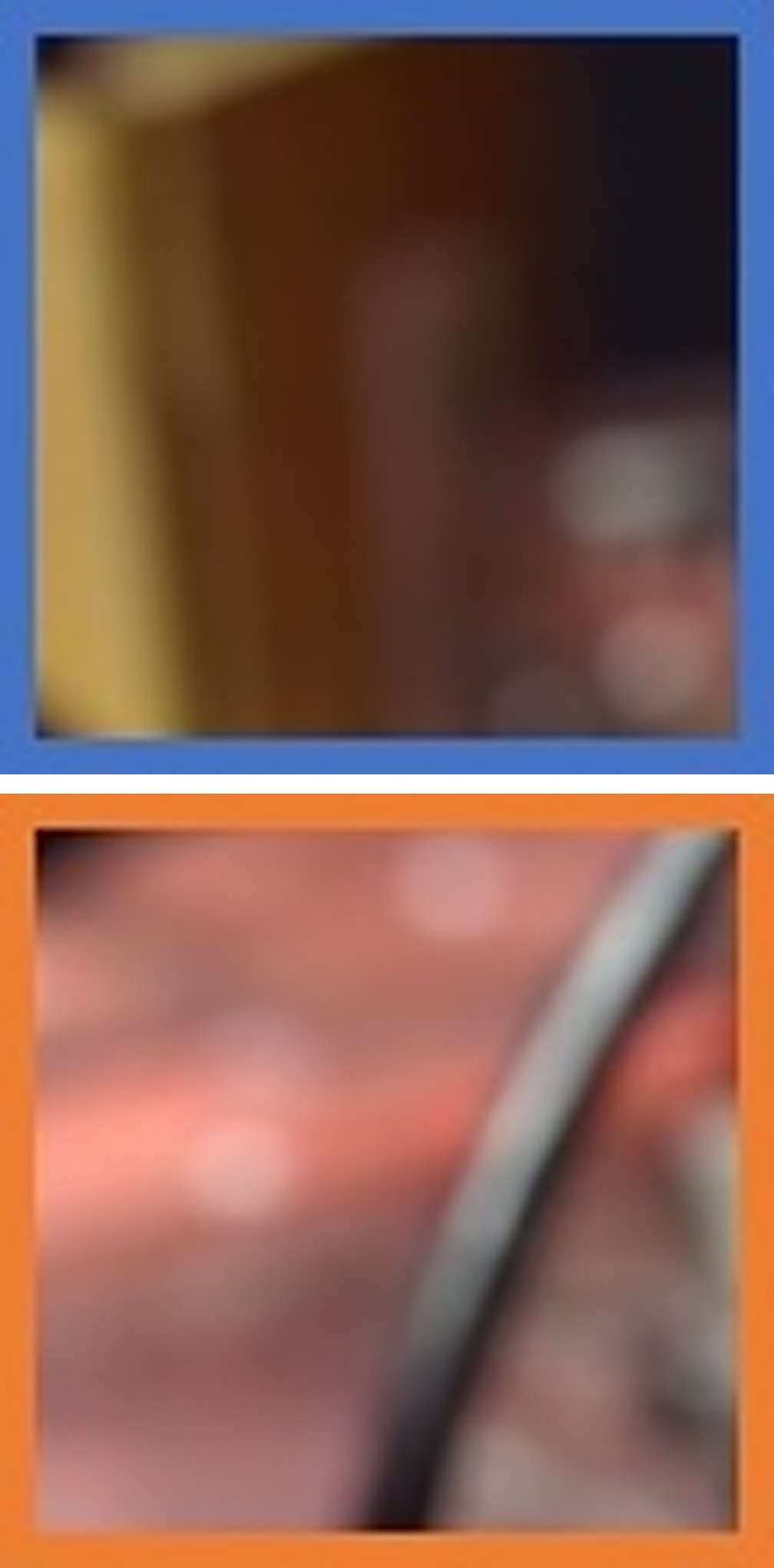} \\

\includegraphics[height=0.38\textwidth]{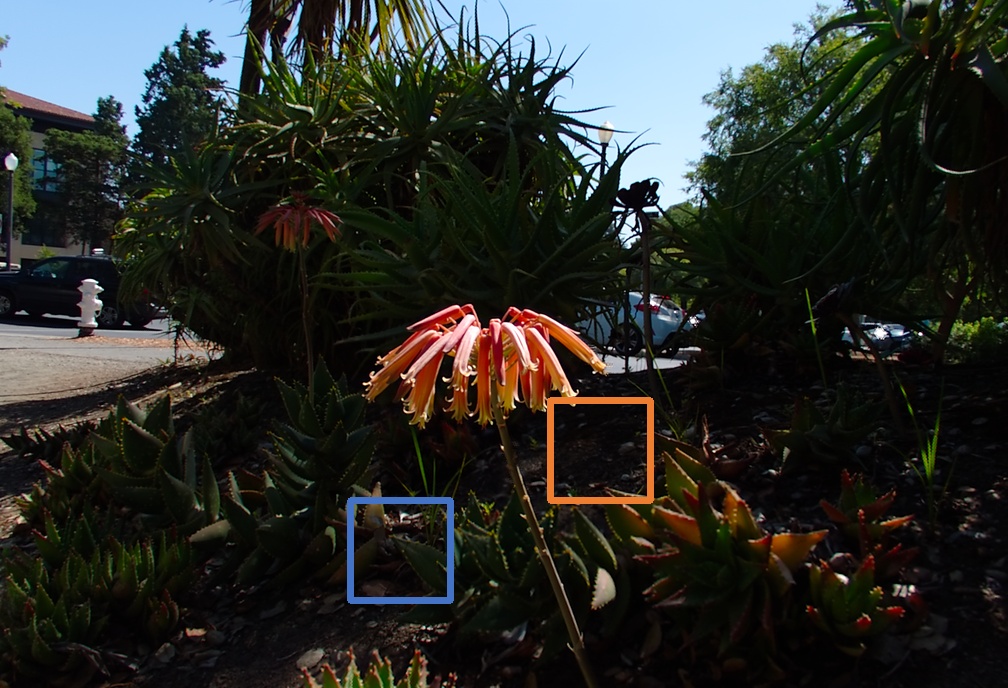} & 
\includegraphics[height=0.38\textwidth]{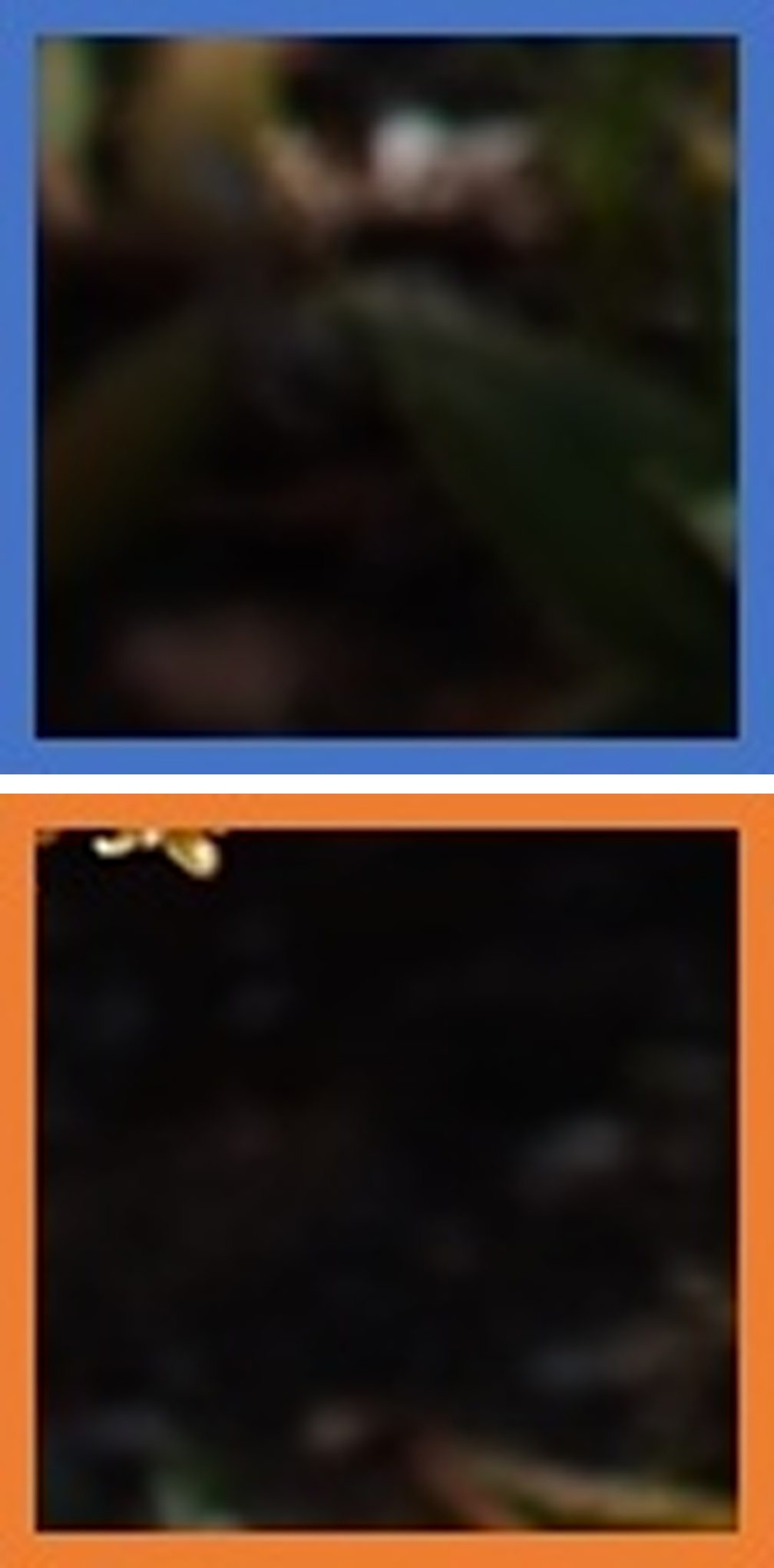} & 
\includegraphics[height=0.38\textwidth]{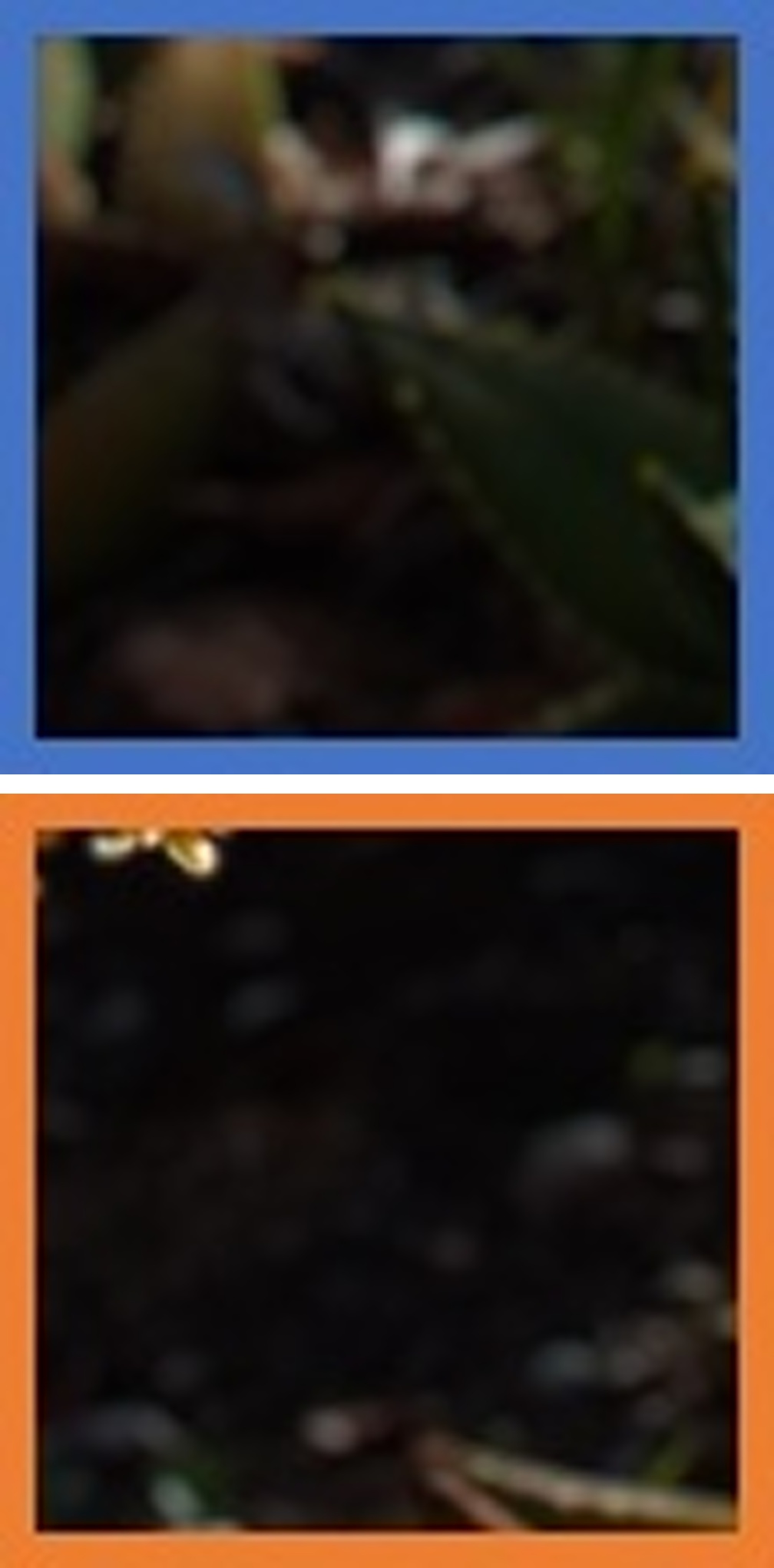} & 
\includegraphics[height=0.38\textwidth]{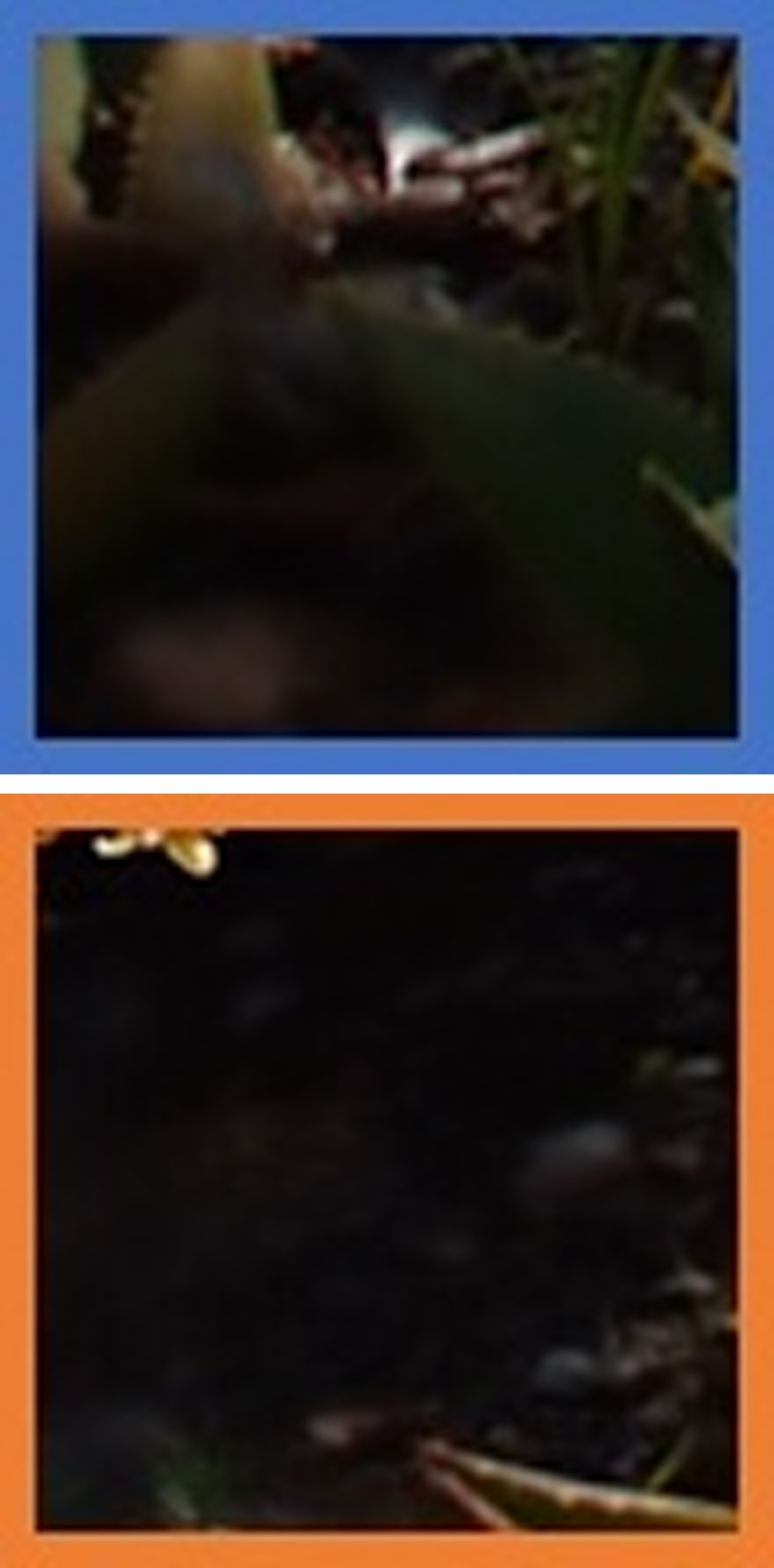} & 
\includegraphics[height=0.38\textwidth]{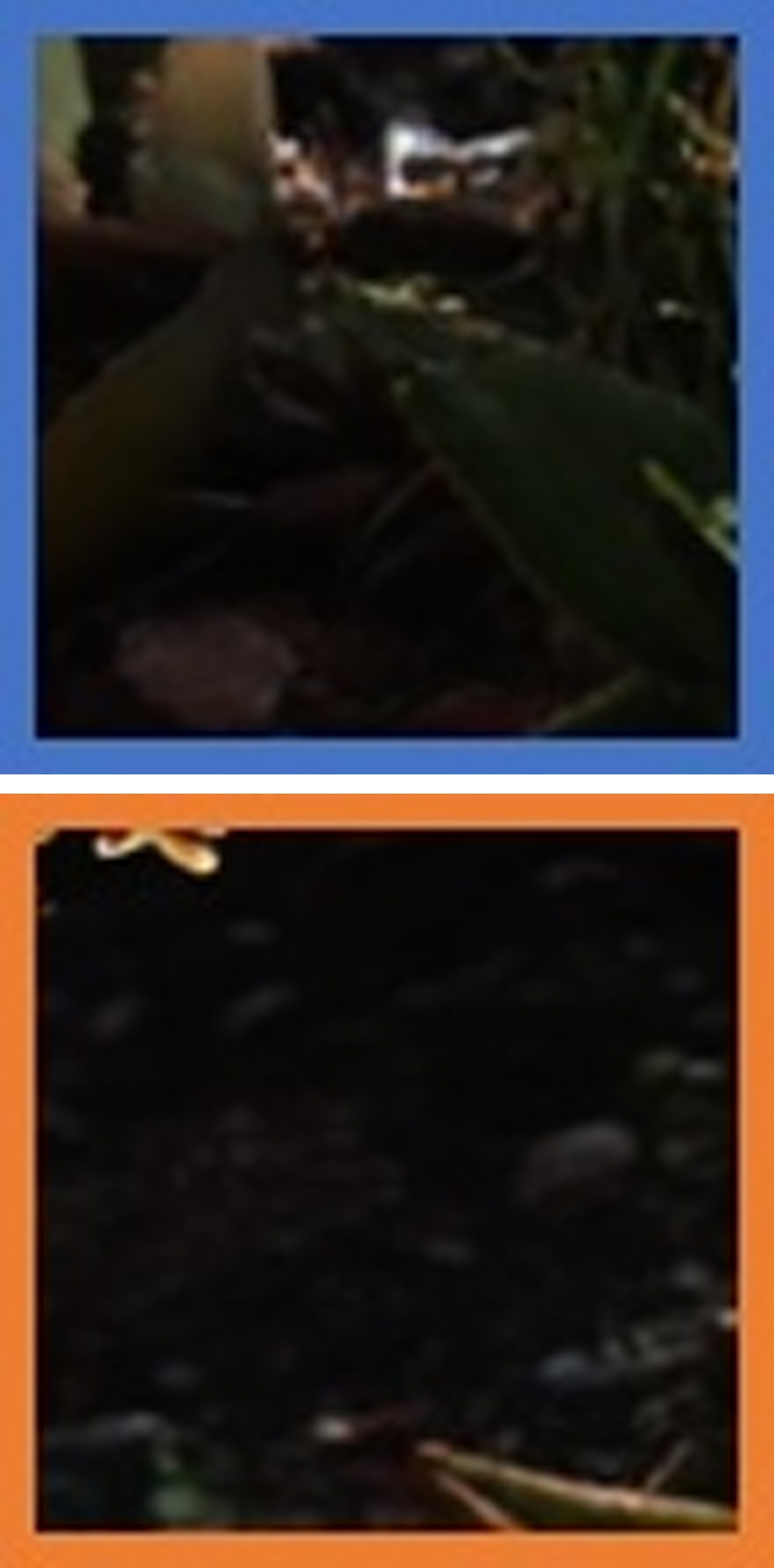} & 
\includegraphics[height=0.38\textwidth]{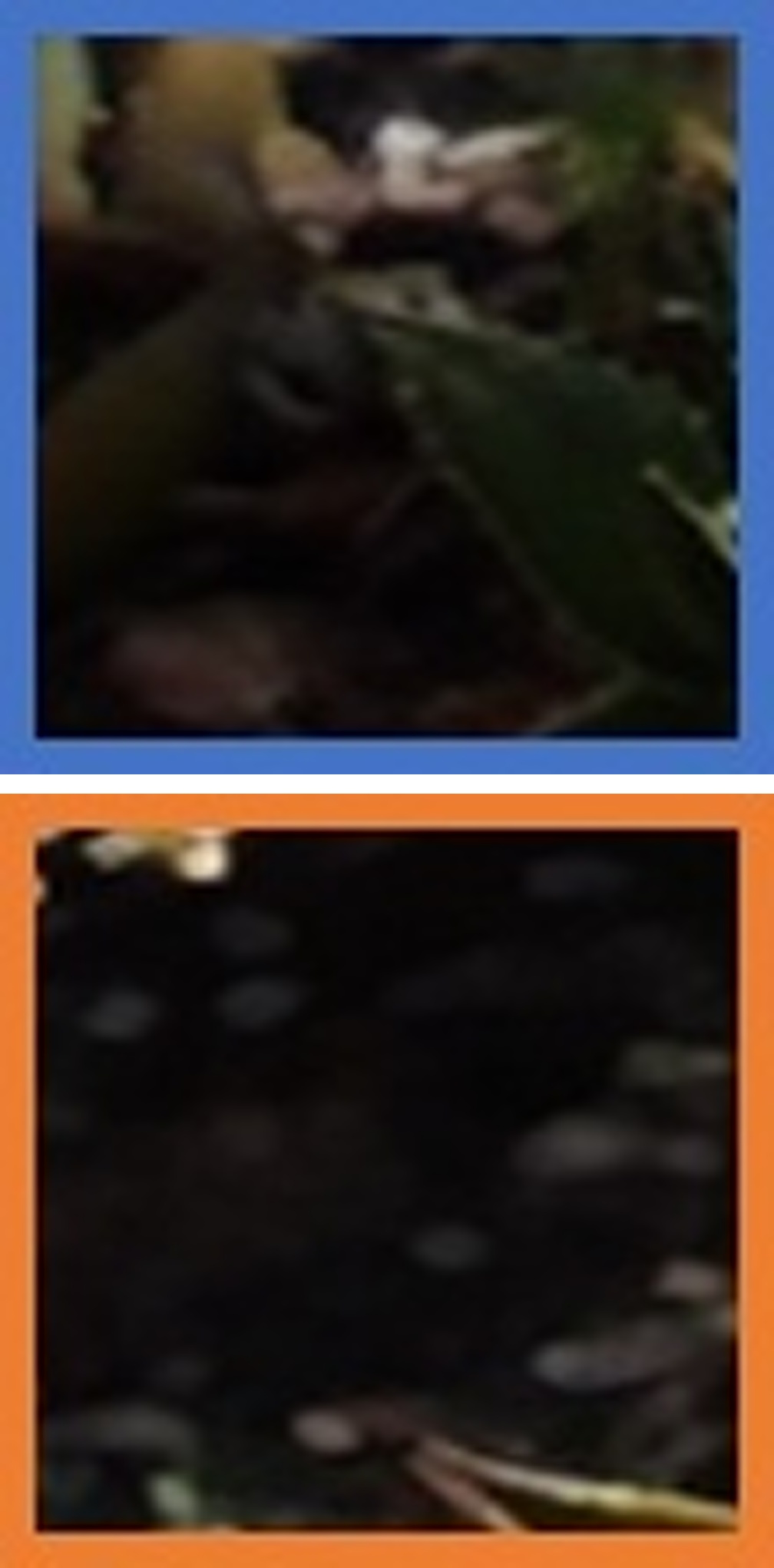} & 
\includegraphics[height=0.38\textwidth]{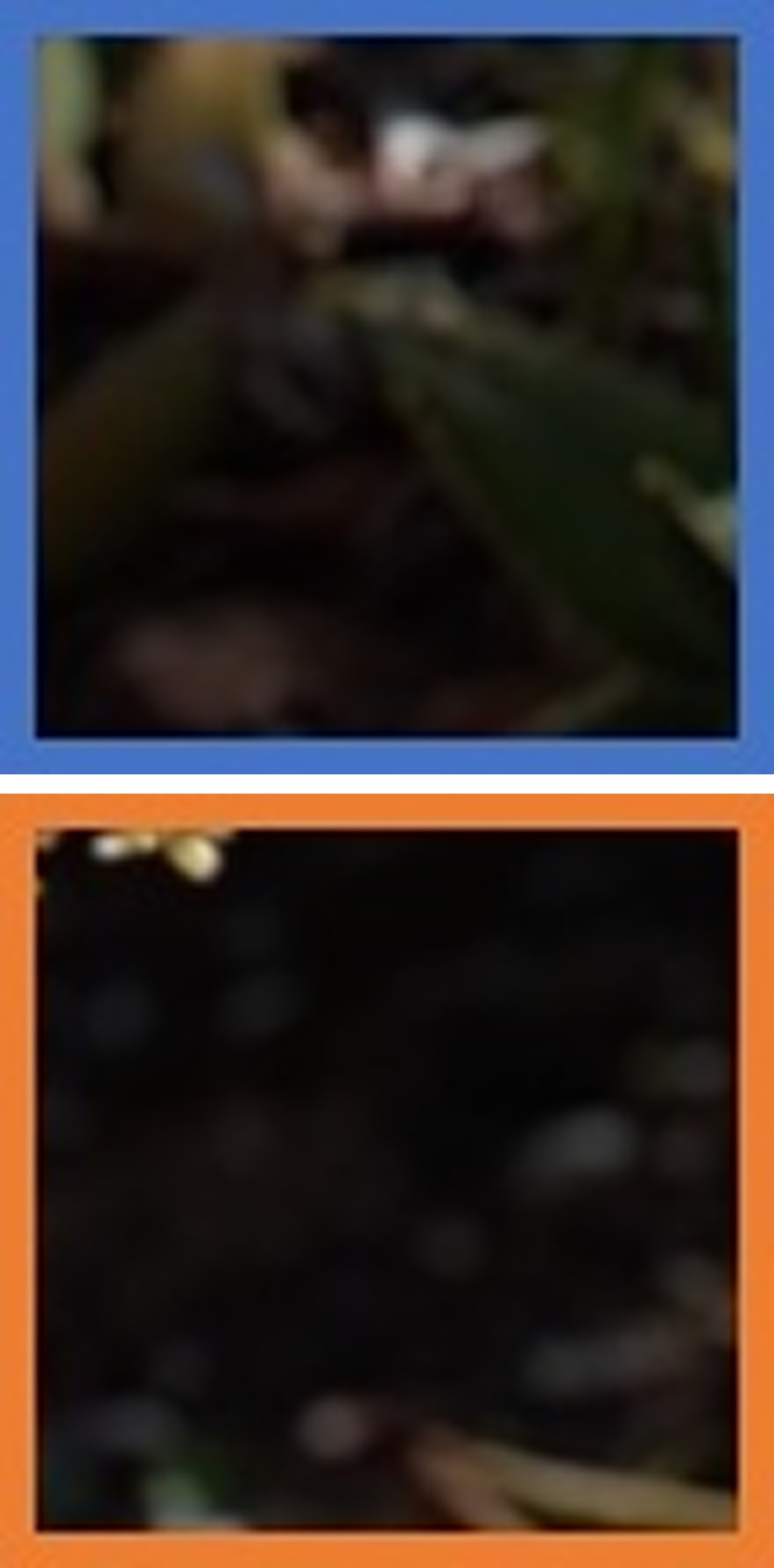} \\
Input & GT & BokehMe & Bokehlicious & BokehDiff & Diffcamera & GenRefocus \\
 &  & \cite{BokehMe} & \cite{Bokehlicious} & \cite{BokehDiff} & \cite{wang2025diffcamera} & (Ours) \\
\end{tabular}
}
\caption{
\textbf{Qualitative comparison on bokeh synthesis benchmark.} Results on LF-Bokeh with zoomed patches (\textcolor{blue}{blue} and \textcolor{orange}{orange} boxes) highlighting detail quality. Our method synthesizes bokeh effects that better match ground truth with realistic blur gradients and natural occlusion handling.
}
\label{fig:bokeh}
\end{figure*}

\subsubsection{Benchmarks and Metrics}
We evaluate our method on three tasks using reference-based fidelity and no-reference image quality assessment metrics.

\paragraph{(i) Defocus Deblurring.} 
We conduct deblurring experiments on RealDOF and DPDD datasets. To assess reconstruction fidelity and perceptual naturalness, we report LPIPS~\cite{LPIPS}, DISTS~\cite{DISTS}, and no-reference metrics including CLIP-IQA~\cite{CLIP_IQA}, MANIQA~\cite{MANIQA}, and MUSIQ~\cite{MUSIQ}.

\paragraph{(ii) Bokeh Synthesis.} 
We introduce LF-Bokeh, featuring $200$ images with diverse focus planes and aperture sizes synthesized from light-field captures~\cite{EPFL, Dansereau2019LIFF}. In real-world scenarios, the absence of ground-truth depth maps and reference bokeh levels $K$ precludes the computation of accurate defocus maps, making the direct evaluation of fidelity and controllability challenging. To assess fidelity, following prior works~\cite{BokehDiff, BokehMe}, we conduct a per-image \emph{binary search} over $K$ and select the value that \emph{maximizes SSIM} with the target, reporting LPIPS, DISTS, and CLIP-I~\cite{CLIP_I}. To quantify controllability under these constraints, we adopt the LVCorr metric following~\cite{yuan2025generative, Fortes2025BokehDiffusion, wang2025diffcamera}. Specifically, for images generated from the same all-in-focus input with a fixed focus plane across varying bokeh levels $K$, we compute the Pearson correlation coefficient (CorrCoef) between these $K$ values and the Laplacian variance trend, thereby reflecting the model's accuracy in rendering the physical blur progression.

\paragraph{(iii) Refocusing.} 
The LF-Refocus dataset consists of $400$ source--target pairs from LF-Bokeh. Similar to bokeh synthesis, we employ an optimal $K$ search for each pair. We report LPIPS, DISTS, and CLIP-I for fidelity, alongside CLIP-IQA and MUSIQ to evaluate perceptual quality across varying depths of field.

\subsection{Comparison}
\label{sec:comparison}

\subsubsection{Defocus deblurring.}
As reported in Tab.~\ref{tab:realdof_dpdd_fullwidth}, DeblurNet outperforms competing methods across most referenced and non-referenced metrics. While our method yields slightly lower reference scores than Bokehlicious\cite{Bokehlicious} on RealDOF\cite{AIFNet}, it consistently produces more visually appealing results and demonstrates greater robustness to diverse input conditions, such as grayscale photographs and low-quality, blurry images. Extensive visual comparison supporting this flexibility is provided in the supplementary material. 
Qualitatively, this perceptual superiority is evident in Fig.~\ref{fig:defocus deblur}. As shown in Fig.~\ref{fig:defocus deblur}(a), our approach achieves faithful recovery of text details with well-preserved edges. Furthermore, the scene in Fig.~\ref{fig:defocus deblur}(b) presents a particularly challenging case: while competing methods recover only a blurry dark stripe and introduce background distortions, DeblurNet reconstructs more consistent content and delivers compelling results.


\begin{table*}[t]
  \centering
  \footnotesize
  \setlength{\tabcolsep}{3pt}
  \caption{
  \textbf{Quantitative comparison on refocusing benchmark.} Results on our LF-Refocus dataset. We compare our method against current state-of-the-art refocusing pipelines~\cite{wang2025diffcamera, tedla2025learning}. Optimal bokeh level $K$ is determined via per-image binary search following~\cite{BokehDiff, BokehMe}.$^\dagger$Learn2Refocus~\cite{tedla2025learning} takes a specific focal position as the input condition to synthesize the remaining images in the focal stack. To ensure coverage of the full range of focal planes, we perform inference twice separately: once conditioned on the minimum focal position (1) and once on the maximum (9). We then evaluate all outputs from both generated stacks and report the one yielding the highest SSIM.
}
  \label{tab:refocus}
  \begin{tabular}{lcccccc}
    \toprule
    {Method} & {LPIPS} $\downarrow$ & {DISTS} $\downarrow$ & {CLIP-I} $\uparrow$ & {CLIP-IQA} $\uparrow$ & {MUSIQ} $\uparrow$ \\
    \midrule
    DiffCamera~\cite{wang2025diffcamera}           & \cellcolor{orange!25}0.1426 & \cellcolor{orange!25}0.0782 & \cellcolor{orange!25}0.9401 & 0.4669 & \cellcolor{orange!25}40.5869\\
    Learn2Refocus$^\dagger$ ~\cite{tedla2025learning}        & 0.1706 &  0.0967 &  0.9322 & \cellcolor{orange!25}0.4709 & 38.3440 \\
    GenRefocus (Ours)         & \cellcolor{red!25}0.1324 & \cellcolor{red!25}0.0759 & \cellcolor{red!25}0.9571 & \cellcolor{red!25}0.5803 & \cellcolor{red!25}52.9706  \\
    \bottomrule
  \end{tabular}
\end{table*}
\begin{figure*}[t]
\centering
\scriptsize
\setlength{\tabcolsep}{2pt}
\resizebox{\textwidth}{!}{
\begin{tabular}{ccccc}
\includegraphics[height=0.32\textwidth]{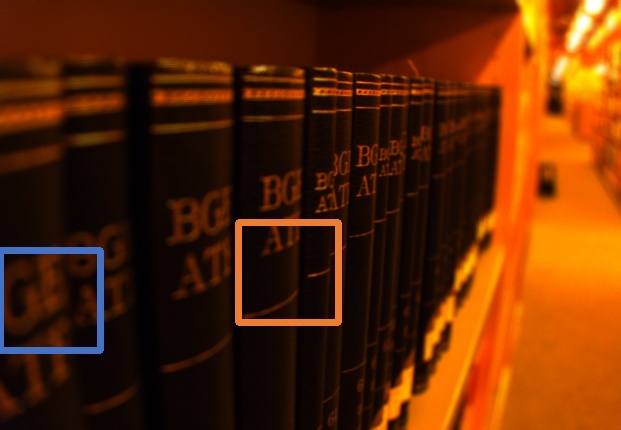} & 
\includegraphics[height=0.32\textwidth]{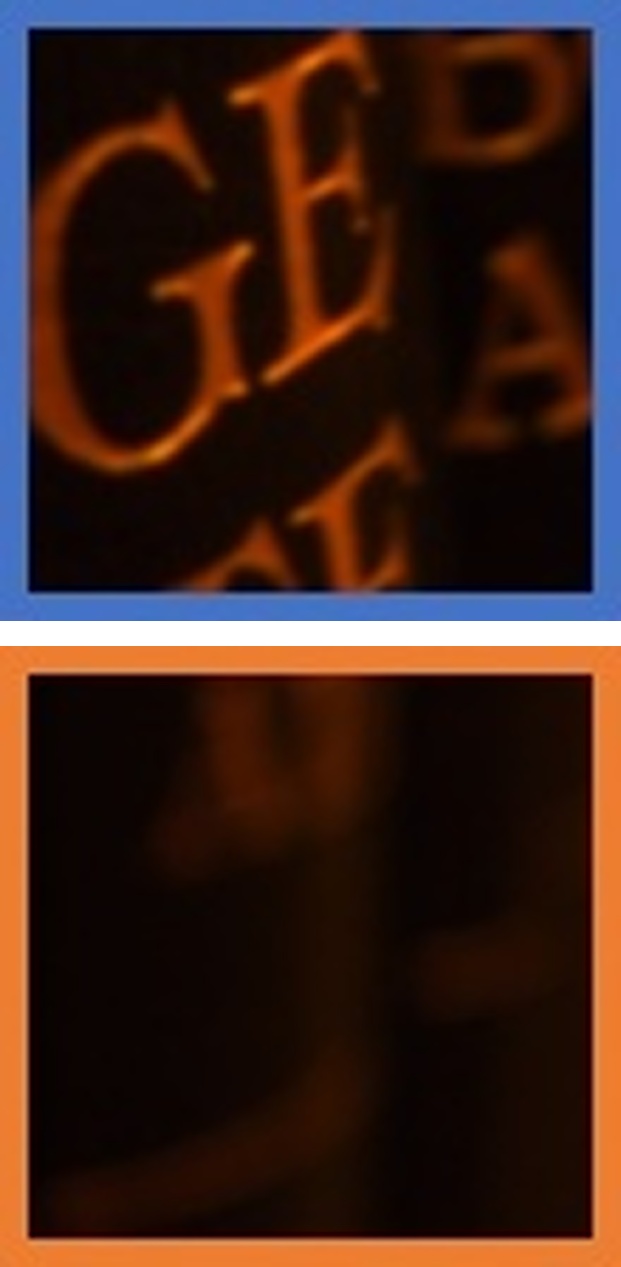} & 
\includegraphics[height=0.32\textwidth]{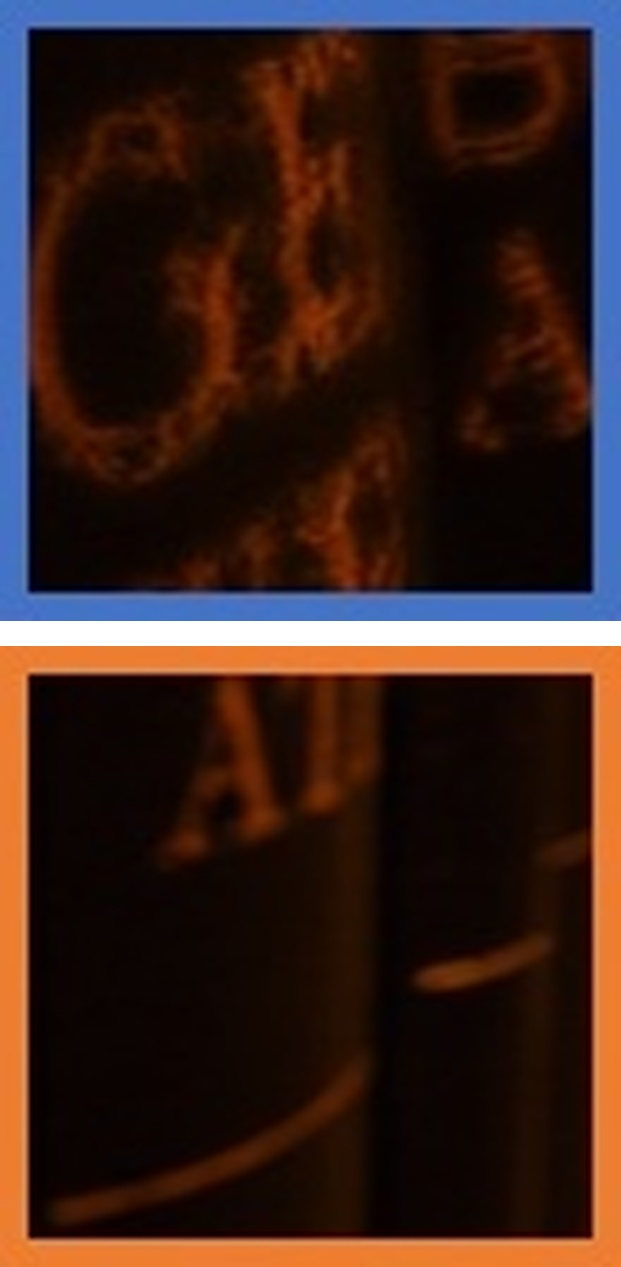} & 
\includegraphics[height=0.32\textwidth]{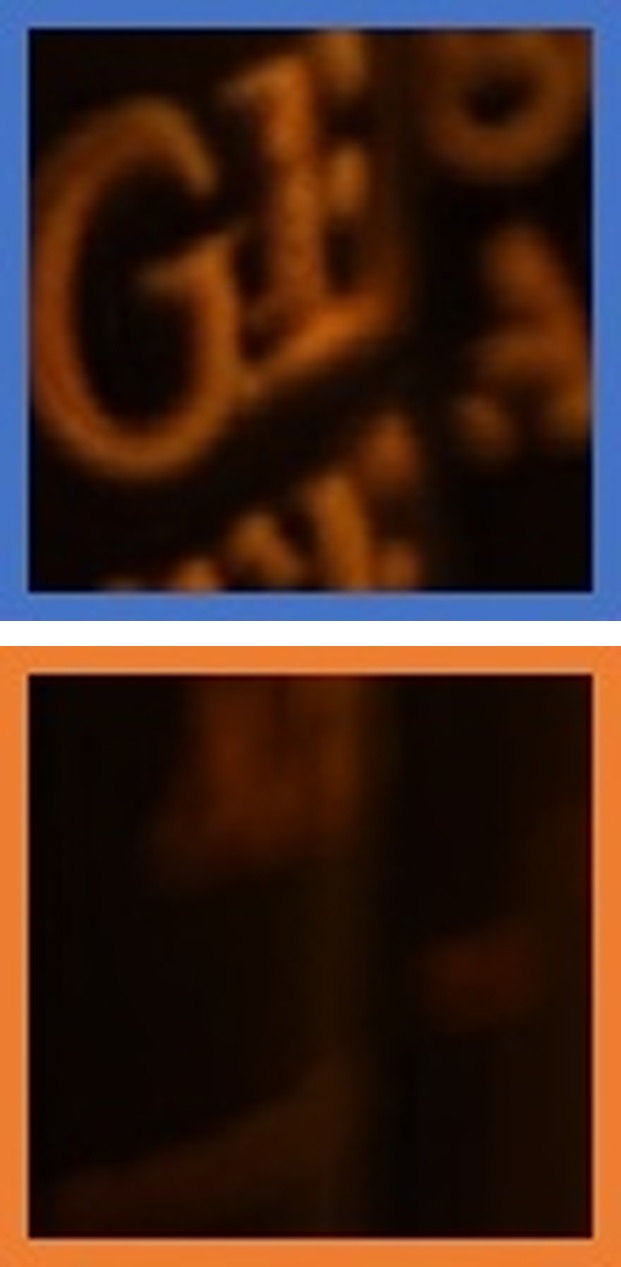} & 
\includegraphics[height=0.32\textwidth]{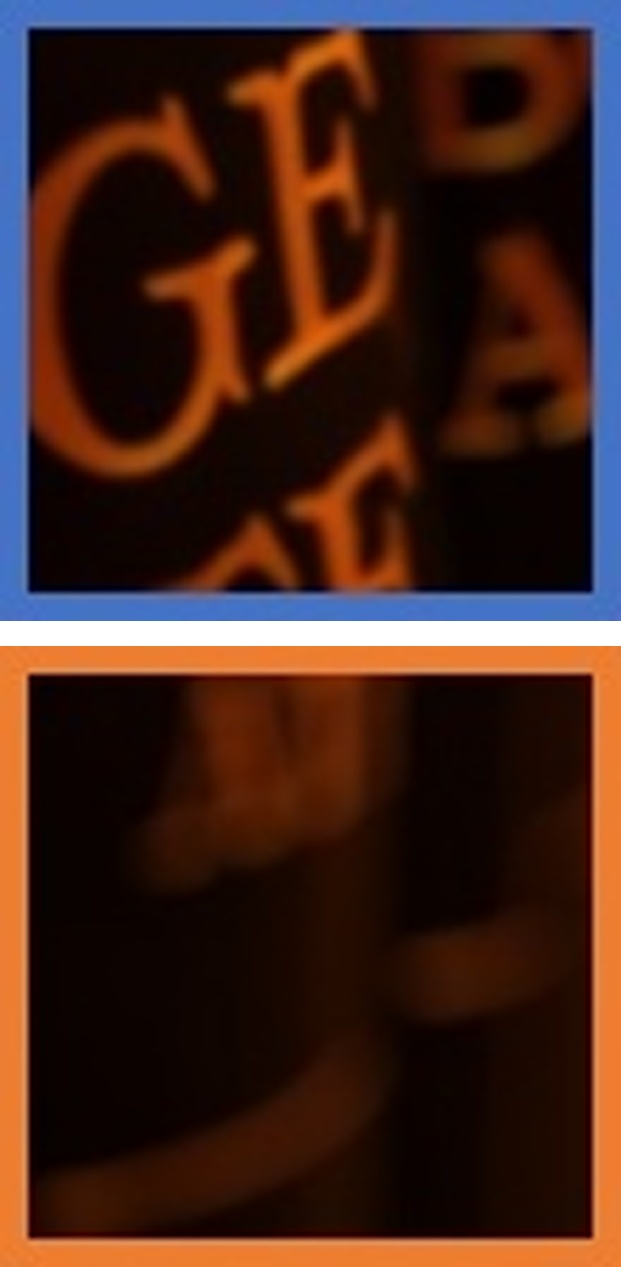} \\

\includegraphics[height=0.32\textwidth]{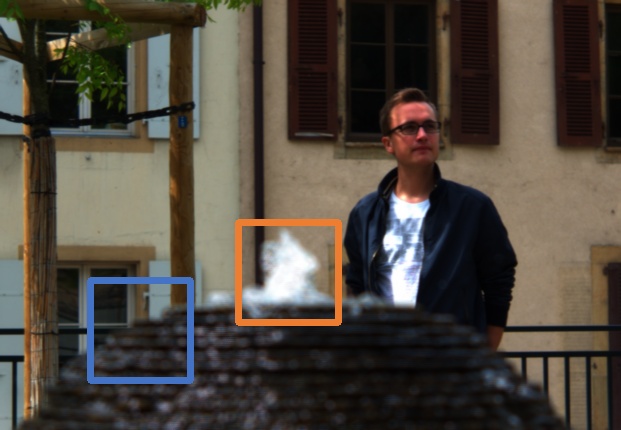} & 
\includegraphics[height=0.32\textwidth]{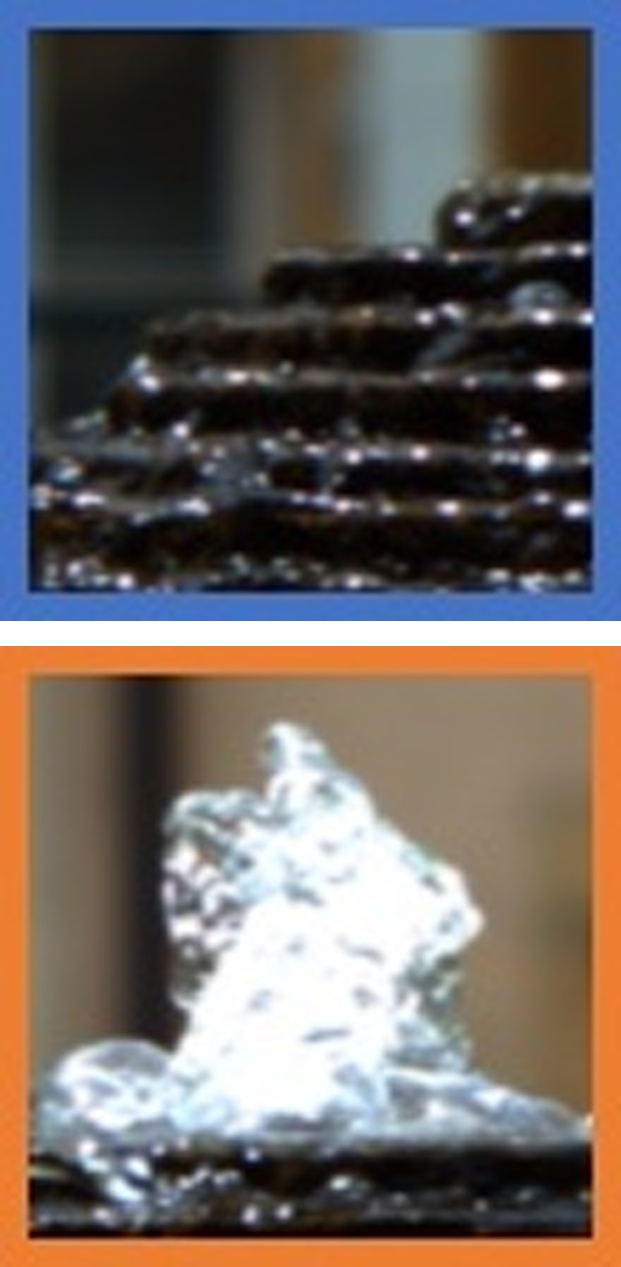} & 
\includegraphics[height=0.32\textwidth]{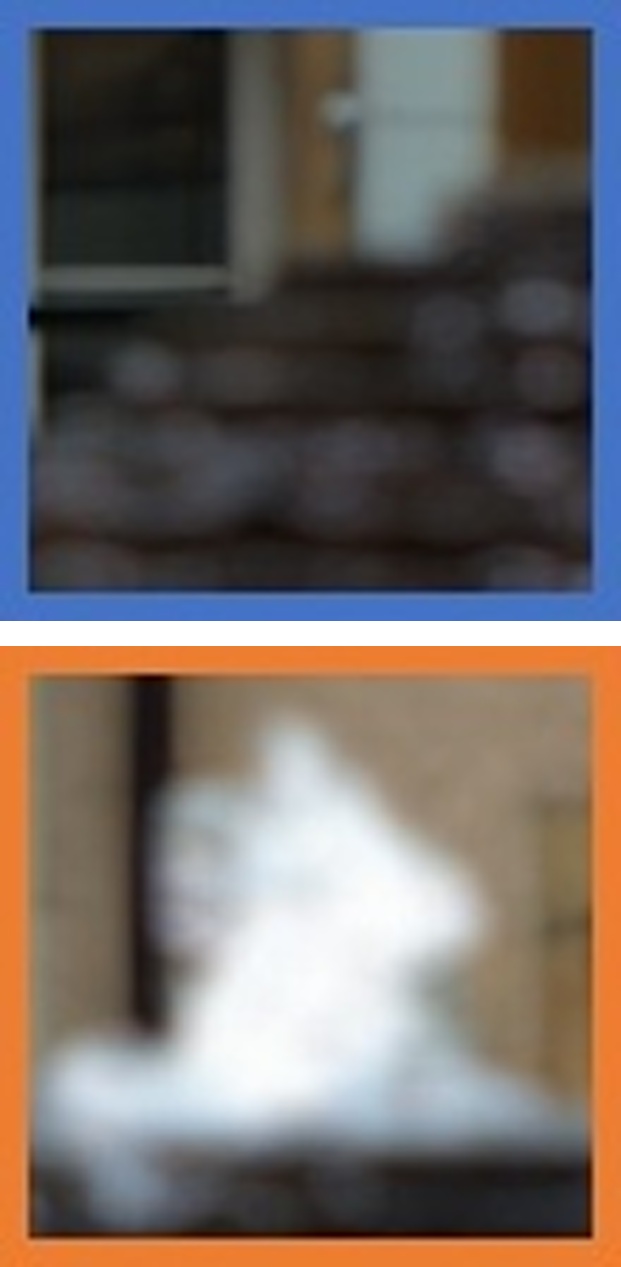} & 
\includegraphics[height=0.32\textwidth]{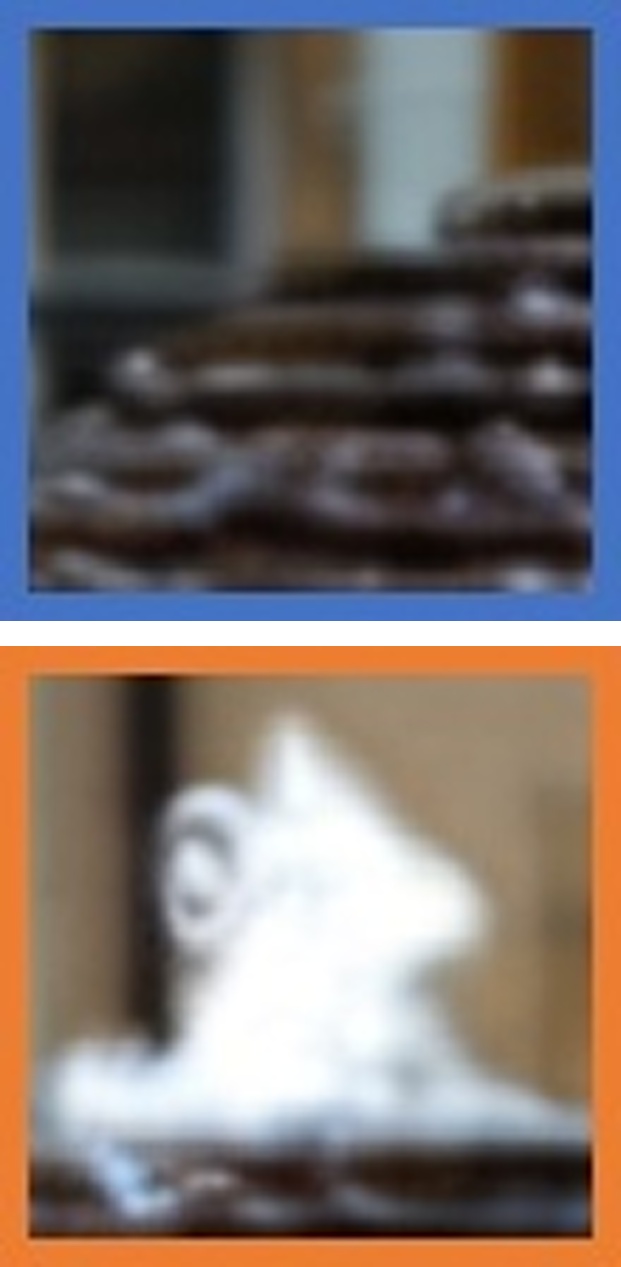} & 
\includegraphics[height=0.32\textwidth]{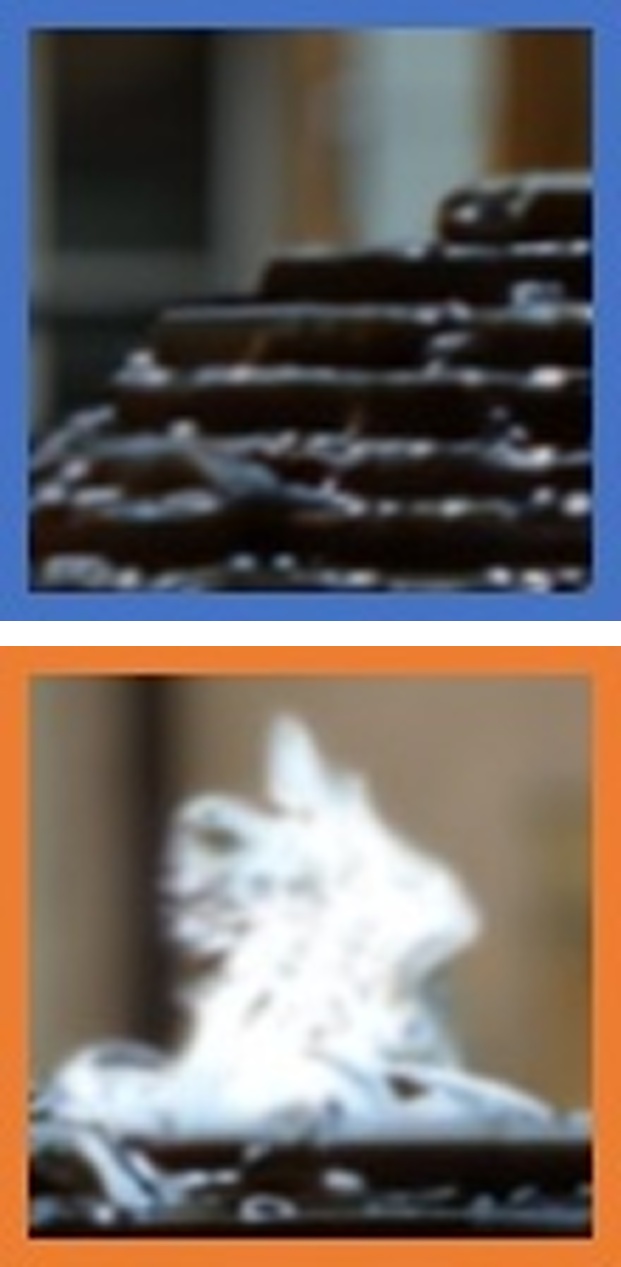} \\
Input & GT & DiffCamera & Learn2Refocus & GenRefocus \\
 &  & \cite{wang2025diffcamera} & \cite{tedla2025learning} & (Ours) \\
\end{tabular}
}
\caption{
\textbf{Qualitative comparison on the refocusing benchmark.} Results on the LF-Refocus dataset with zoomed patches (\textcolor{blue}{blue} and \textcolor{orange}{orange} boxes) highlight the fidelity of focal and defocused regions. We observe that DiffCamera~\cite{wang2025diffcamera} struggles to reconstruct sharp structural details due to its strict input conditions and often assigns incorrect blurriness levels to background regions. Similarly, Learn2Refocus~\cite{tedla2025learning} fails to achieve accurate focus-plane placement, resulting in unintended blur of the target object. In contrast, our method achieves precise refocusing at the target plane while synthesizing photorealistic bokeh, closely matching the ground truth across diverse scenes.
}
\label{fig:refocus_depth_visual}
\end{figure*}

\subsubsection{Bokeh synthesis.}
As presented in Tab.~\ref{tab:bokeh}, our approach outperforms all other methods across fidelity metrics. In terms of controllability, which measures how well the generated bokeh aligns with the conditioned bokeh level $K$, BokehMe~\cite{BokehMe} achieves near-perfect control owing to its hybrid design that incorporates a classical, physically motivated renderer. 
Nevertheless, our method outperforms all diffusion-based baselines, demonstrating that our learning strategy effectively preserves both realism and accuracy. Furthermore, it is worth noting the specific limitations of competing methods: DiffCamera~\cite{wang2025diffcamera} requires inputs with a fixed resolution and aspect ratio due to its reliance on explicit focus point conditioning, rendering it incompatible with tiling strategies; consequently, resizing operations severely degrade its controllability. Meanwhile, Bokeh Diffusion~\cite{Fortes2025BokehDiffusion} relies on an inversion~\cite{song2020denoising} process that not only prolongs inference time but also strips the image of its original high-frequency details, thereby lowering overall fidelity. The qualitative results shown in Fig.~\ref{fig:bokeh}.

\subsubsection{Refocusing.}
As shown in Tab.\ref{tab:refocus}, our approach achieves the best results on both fidelity and perceptual metrics, outperforming other baselines. 
Qualitatively, Fig.~\ref{fig:refocus_depth_visual} shows that competing methods tend to produce spatially ambiguous blur and fail to place the focal plane correctly, resulting in uniformly soft content.
In contrast, our approach restores fine details at the intended focus while synthesizing physically consistent bokeh elsewhere.

\subsection{Ablation Study}
\label{sec:ablation}

\begin{table}[t]
  \centering
  \footnotesize
  \setlength{\tabcolsep}{4pt}
  \caption{
  \textbf{Comparison of pipeline designs.} 
  Comparing direct refocusing with our proposed deblurring, then the bokeh synthesis approach on LF-Refocus. 
  }
  \label{tab:pipeline-comparison} 
  \begin{tabular}{lccc}
    \toprule
    Method  & LPIPS~$\downarrow$ & DISTS~$\downarrow$ & CLIP-I~$\uparrow$\\
    \midrule
    Direct refocusing  & 0.1723 & 0.0976 & 0.9345  \\
    Deblurring then bokeh synthesis (Ours) & \textbf{0.1324} & \textbf{0.0759} & \textbf{0.9571} \\
    \bottomrule
  \end{tabular}
\end{table}

\begin{table}[t]
  \centering
  \footnotesize
  \setlength{\tabcolsep}{3pt}
  \caption{
  \textbf{Ablation on BokehNet training datasets.} 
  Comparing combinations of datasets: (a) Synthetic data, (b) ITW dataset\cite{Fortes2025BokehDiffusion}, and (c) LFDOF~\cite{AIFNet} and RealBokeh\cite{Bokehlicious} on LF-Bokeh. 
  The results show that combining all three yields the best optical characteristics.
  }
  \label{tab:ablation-train}
  \begin{tabular}{cccccc}
    \toprule
    \multicolumn{3}{c}{Datasets} & \multicolumn{3}{c}{Metrics} \\
    \cmidrule(lr){1-3} \cmidrule(lr){4-6}
    (a) Synthetic & (b) ITW & (c) LFDOF \& RealBokeh & LPIPS~$\downarrow$ & DISTS~$\downarrow$ & CLIP-I~$\uparrow$ \\
    \midrule
    \checkmark & - & - & 0.1289 & 0.0738 & 0.9461 \\
    \checkmark & \checkmark & - & 0.1156 & 0.0665 & 0.9529 \\
    \checkmark & - & \checkmark & \cellcolor{orange!25}0.0972 & \cellcolor{orange!25}0.0560 & \cellcolor{orange!25}0.9666 \\
    \checkmark & \checkmark & \checkmark & \cellcolor{red!25}\textbf{0.0833} & \cellcolor{red!25}\textbf{0.0487} & \cellcolor{red!25}\textbf{0.9713} \\
    \bottomrule
  \end{tabular}
\end{table}

\subsubsection{Direct refocusing vs. deblurring then bokeh synthesis.}
We evaluate a baseline that \emph{directly} predicts a refocused image from a blurry input and a defocus map. Trained on identical synthetic data using the BokehNet backbone, this direct approach significantly underperforms our two-stage pipeline (Tab.~\ref{tab:pipeline-comparison}). We attribute this gap to two factors: (1) \emph{control fidelity}: estimating depth from blurry inputs degrades defocus map accuracy, causing misaligned blur fields; and (2) \emph{data leverage}: a direct formulation precludes injecting tailored real-data supervision into separate deblurring and synthesis subtasks.

\subsubsection{BokehNet training ablation.}
To quantify the impact of different training data distributions, we evaluate BokehNet training under various dataset combinations on LF-Bokeh. We consider three primary data sources: (a) synthetic data, (b) ITW dataset\cite{Fortes2025BokehDiffusion}, and (c) LFDOF~\cite{AIFNet} and RealBokeh~\cite{Bokehlicious}.
As shown in Tab.~\ref{tab:ablation-train}, a baseline trained purely on (a) provides a foundational performance but leaves room for improvement. Introducing either (b) or (c) to the synthetic baseline noticeably enhances performance. Ultimately, the comprehensive setting that combines all three sources yields the most substantial improvements, delivering the best optical characteristics and real-world generalization.


\subsubsection{Aperture shape control.} 
\begin{figure*}[t]
  \centering
  \footnotesize
  \setlength{\tabcolsep}{1pt}
  \resizebox{\linewidth}{!}{
  \begin{tabular}{cccc}
  \includegraphics[width=0.25\linewidth]{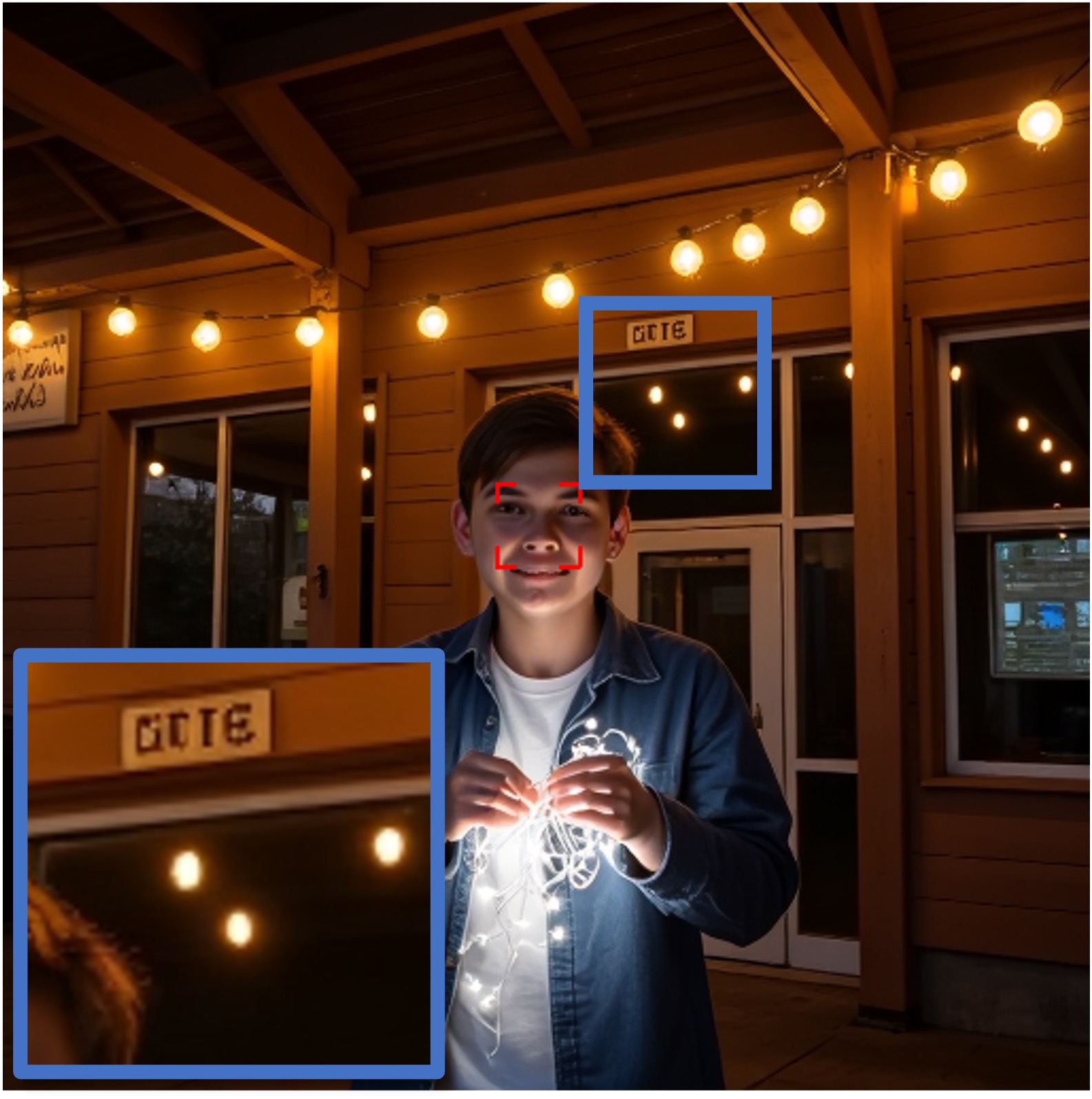} & 
  \includegraphics[width=0.25\linewidth]{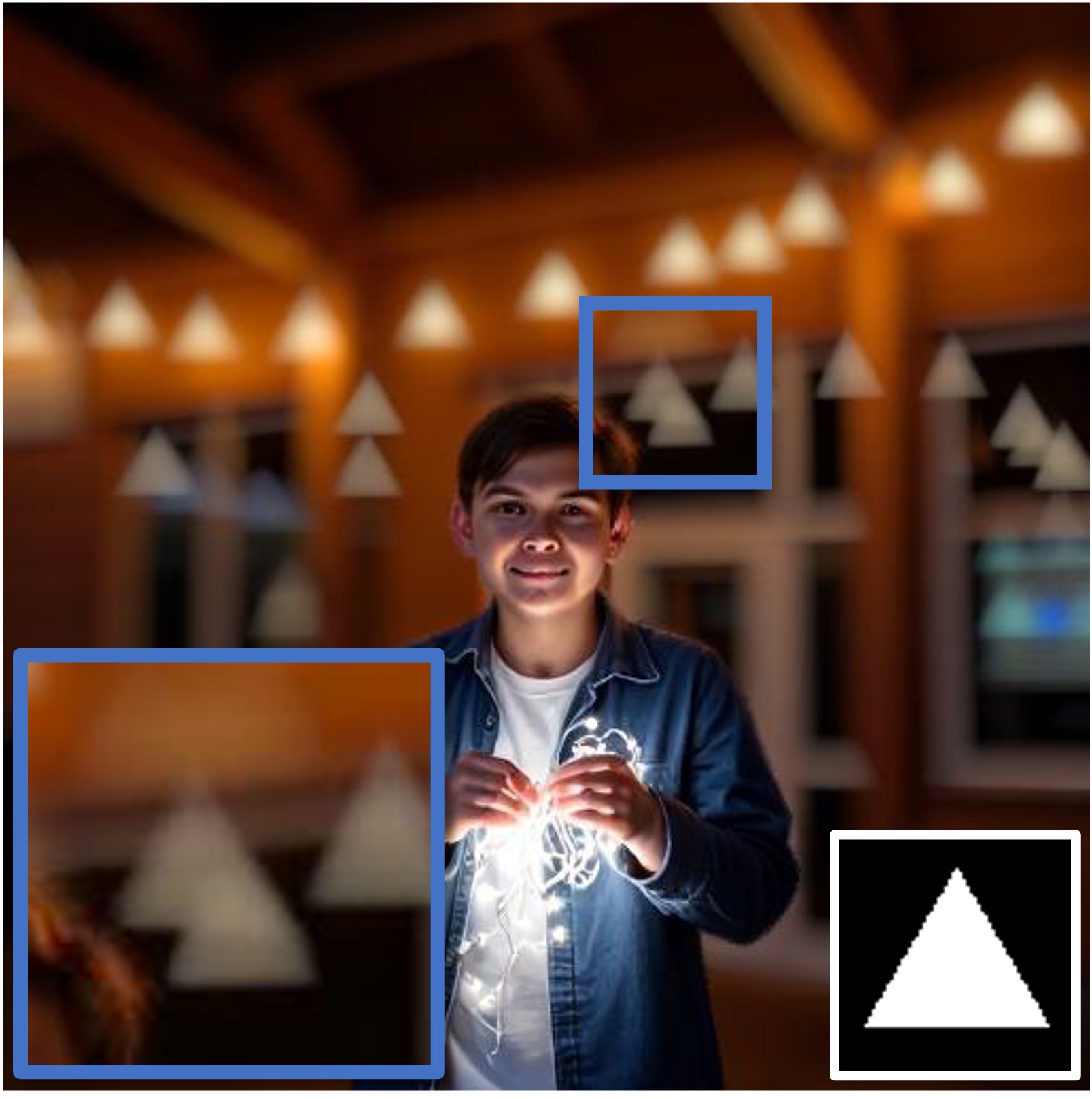} & 
  \includegraphics[width=0.25\linewidth]{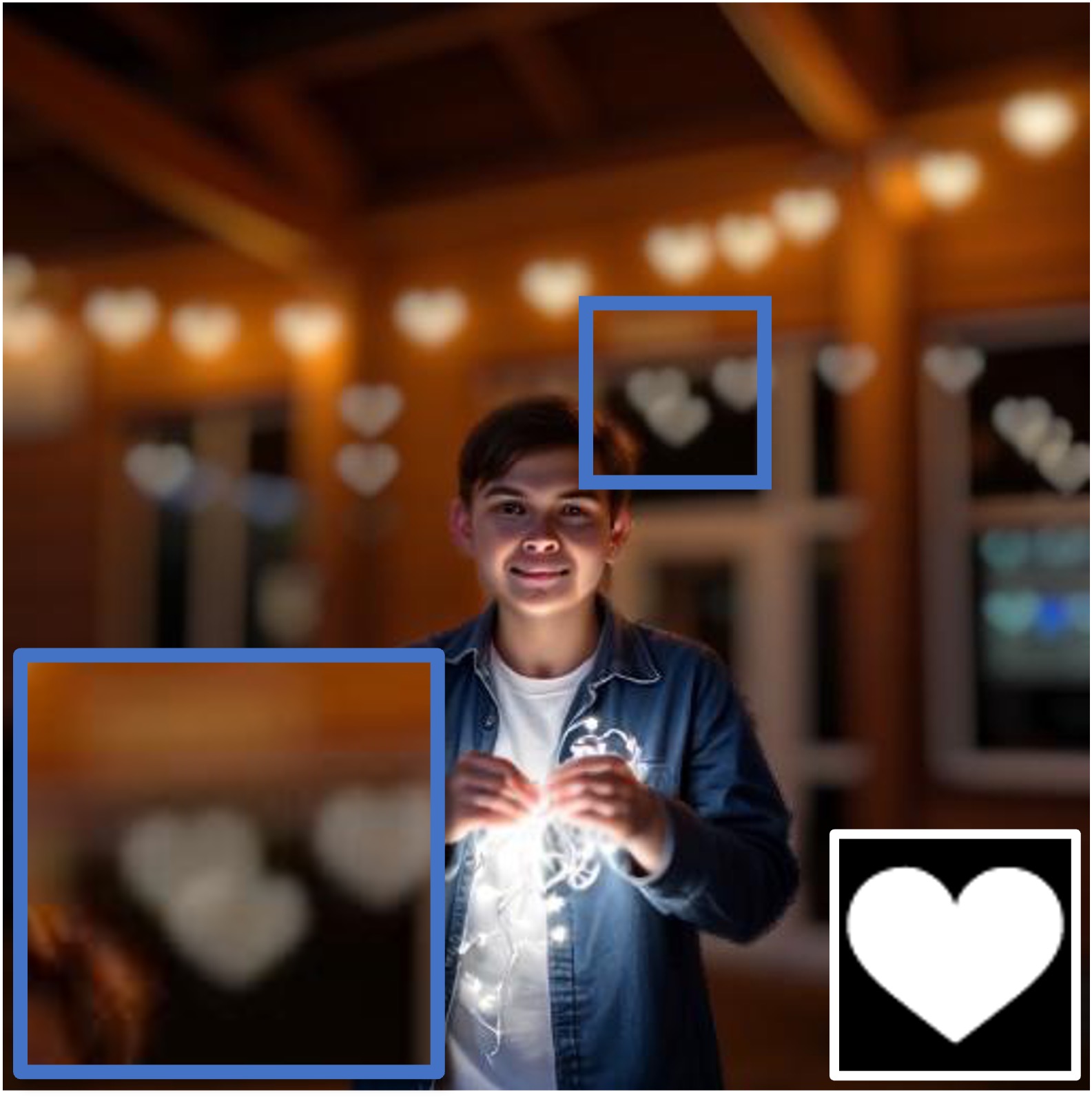} & 
  \includegraphics[width=0.25\linewidth]{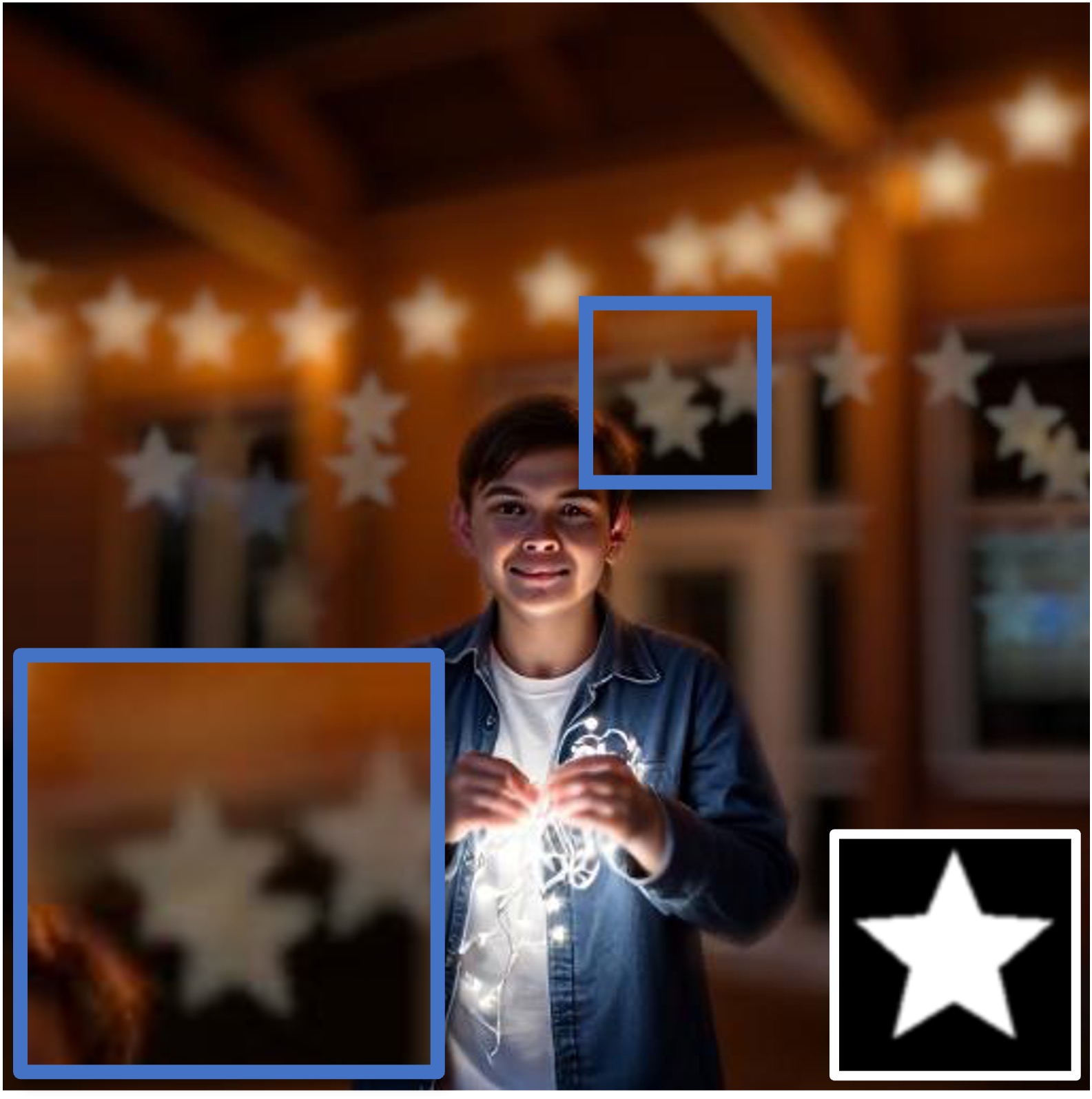} \\
  Input & Triangle & Heart & Star \\
  \end{tabular}
  }
  \caption{
  \textbf{Controllable aperture shape synthesis.} Given an input image, our BokehNet can synthesize bokeh effects with custom aperture shapes (triangle, heart, star) by conditioning on shape exemplars. Background point lights exhibit the specified aperture geometry while maintaining scene consistency. 
  }
  \label{fig:bokeh shape}
\end{figure*}

As illustrated in Fig.~\ref{fig:bokeh shape}, after we fine-tune BokehNet on our point-light 1K dataset, the model can synthesize bokeh with diverse, user-specified aperture shapes.

\begin{figure*}[t]
  \centering
  \begin{minipage}[t]{0.51\textwidth}
    \centering
    \includegraphics[width=\linewidth]{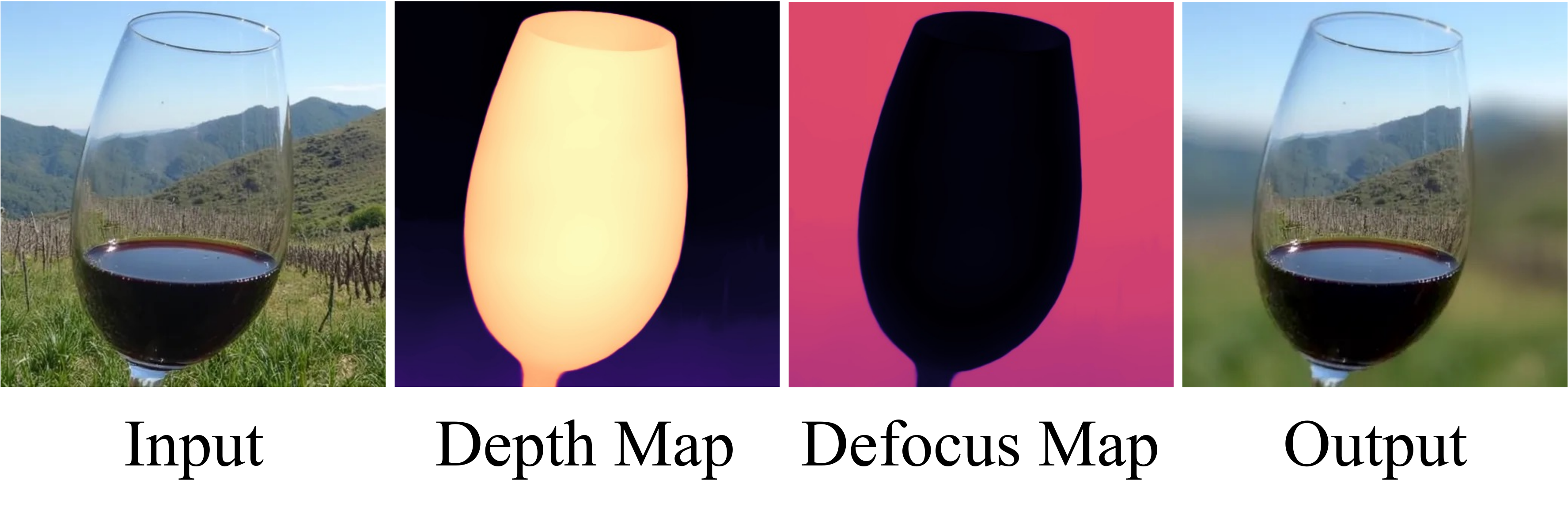}
    \caption{
    \textbf{Failure case: transparent surfaces.}
    With a transparent glass, focusing on the foreground may fail to correctly blur the background due to unreliable defocus maps.
    }
    \label{fig:refraction}
  \end{minipage}
  \hfill 
  \begin{minipage}[t]{0.47\textwidth}
    \centering
    \includegraphics[width=\linewidth]{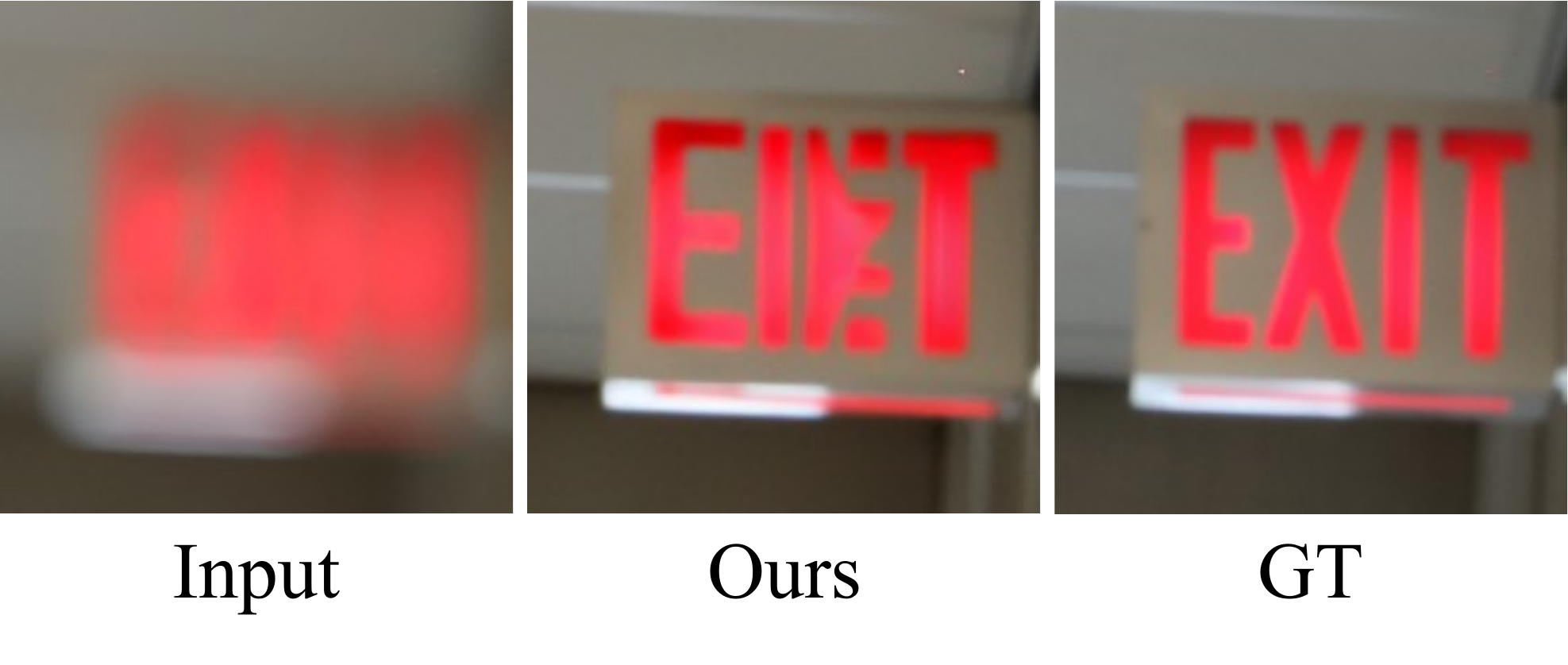}
    \caption{
    \textbf{Failure case: hallucinated details.}
    With severely blurred inputs, our model may hallucinate incorrect results.
    }
    \label{fig:hallucinate}
  \end{minipage}
\end{figure*}




\section{Conclusion}
We present \emph{Generative Refocusing}, a two-stage diffusion framework turning single images into controllable virtual cameras. By decoupling into DeblurNet (deblurring) and BokehNet (bokeh synthesis), our approach handles arbitrary inputs with intuitive controls over focus plane, bokeh intensity, and aperture shape. Through training on various data sources, GenRefocus learns authentic optical behavior beyond simulators. Experiments on all benchmarks show consistent improvements, enabling aperture-shape editing.

\subsubsection{Limitations and future work.}
GenRefocus relies on the defocus map; Imperfections in the depth estimate can propagate to the defocus map and yield incorrect blur assignment. Transparent surfaces are particularly challenging, as monocular depth typically captures the depth of the front transparent surface while ignoring the background seen through it (see Fig.~\ref{fig:refraction}).
In cases of severe blur, our diffusion-based method may fail to recover fine details and instead hallucinate textures (see Fig.~\ref{fig:hallucinate}), rendering it unsuitable for precision-critical fields such as scientific imaging or safety-sensitive applications. Complex aperture shapes require simulator-driven training data. Future work should generalize to richer aperture vocabularies, including user-drawn designs.


%
%
\bibliographystyle{splncs04}
\bibliography{main}

\clearpage
\appendix
\section{Overview}
In the supplementary material, we provide additional details regarding the training datasets used in our framework and the construction pipeline of PointLight-1K dataset. We also present extended comparisons with DiffCamera~\cite{wang2025diffcamera}, Bokehlicious~\cite{Bokehlicious}, and Vision-Language Models. Furthermore, we provide qualitative results demonstrating the controllability of our approach, along with further analysis on ITW dataset~\cite{Fortes2025BokehDiffusion} and a demonstration of the pre-deblur module's effectiveness. We also present additional in-the-wild results on the accompanying HTML page.

\section{Additional Training Details}
In this section, we detail how we process datasets for training \emph{DeblurNet} and \emph{BokehNet}.

\subsection{DeblurNet Training Data}
\emph{DeblurNet} is trained using a combination of paired defocus-blur data and high-quality sharp images. Specifically, we adopt all images from the official training split of the \textsc{DPDD} \cite{DPDD} dataset, which provides supervised pairs of defocused and all-in-focus images. To supplement this, we incorporate data from \textsc{RealBokeh\_3MP} \cite{Bokehlicious}. Since this dataset contains images with varying degrees of focus, we apply a quality filter by computing the Laplacian variance of each image as a focus measure. By ranking the dataset based on this metric, we retain the top 3,000 sharpest images to serve as additional supervision for the deblurring task.

\subsection{BokehNet Training Data}
Training \emph{BokehNet} requires diverse, high-resolution all-in-focus inputs to simulate realistic bokeh effects. We draw candidate images from \cite{yuan2025generative} and the \textsc{EBB} \cite{ebb} collections, and similar to the \emph{DeblurNet} stage, we use Laplacian variance to filter out blurry examples. This yields a refined pool of approximately 1.7K sharp images. Furthermore, we utilize a comprehensive dataset of 26K real bokeh images. This collection comprises 13K previously filtered and verified images from the \textsc{ITW} dataset \cite{Fortes2025BokehDiffusion}, alongside 13K images newly curated for this work. The newly collected data consists of focus-consistent series captured with varying apertures, containing 2 to 4 images per set. To establish accurate ground truth for these series, we first conducted a manual verification and refinement process on the in-focus masks. This annotation step required 4 to 8 seconds per image, amounting to approximately 8 hours of manual effort in total. Utilizing these masks, we then optimized the parameter $K$ using simulator~\cite{BokehMe}. After obtaining the optimized $K$ values, we applied a Structural Similarity (SSIM) threshold to filter out sub-optimal results, ensuring that only reliable pairs are used for supervision.

\section{PointLight-1K Collection Pipeline}
We construct the \textsc{PointLight-1K} dataset to better model scenes with strong point-light bokeh. The pipeline consists of the following steps:

\begin{enumerate}
  \item \textbf{Keyword mining.}
    We first mine keywords from photographs on Flickr\footnote{\url{https://www.flickr.com/}} 
    that are tagged with terms such as ``night'' or ``bokeh''. These keywords capture typical
    compositions and scene elements with prominent point lights.

  \item \textbf{Prompt expansion.}
  The collected keywords are then expanded into diverse, natural-language prompts using GPT-4o \cite{openai2024gpt4o}. This step enriches the textual descriptions while preserving point-light characteristics.

  \item \textbf{Image generation with fine-tuned FLUX.}
  We use a fine-tuned version of FLUX.1-Dev \cite{ShakkerLabs2024FluxControlNetUnionPro} (FLUX.1-dev-LoRA-AntiBlur LoRA) to generate images from the prompts. The fine-tuning is designed to encourage sharper scenes.

  \item \textbf{Deblurring with DeblurNet.}
  Despite fine-tuning, we observe that images involving point lights may still exhibit mild residual blur. To further reduce this blur, we pass all generated images through our \emph{DeblurNet} module, optimizing each image toward an all-in-focus appearance.

  \item \textbf{Final dataset.}
  After this process, we obtain a curated set of $1$k night-time images with prominent point lights and minimal residual blur. We refer to this dataset as \textsc{PointLight-1K} and use it as a specialized training set to pretrain and refine our learning strategy for controllable bokeh shape.
\end{enumerate}

\begin{table}[t]
\small
\centering
\caption{\textbf{Quantitative comparison on the DPDD dataset using $512 \times 512$ central crops for the defocus deblurring task.} Both methods are evaluated without any tiling strategy to ensure a strictly fair comparison of core architectural capacity. The best results are highlighted in \textbf{bold}.}
\label{tab:supp_diffcamera}
\begin{tabular}{lccccc}
\toprule
Method & LPIPS $\downarrow$ & DISTS $\downarrow$ & CLIP-IQA $\uparrow$ & MANIQA $\uparrow$ & MUSIQ $\uparrow$ \\
\midrule
DiffCamera~\cite{wang2025diffcamera} & 0.2833 & 0.1793 & 0.4602 & 0.3502 & 43.2997 \\
\textbf{Ours} & \textbf{0.1285} & \textbf{0.1045} & \textbf{0.4637} & \textbf{0.3765} & \textbf{47.2070} \\
\bottomrule
\end{tabular}
\end{table}

\section{Additional Comparison with DiffCamera}

To ensure a strictly fair comparison of core model performance, we evaluate both methods under identical input conditions. Because DiffCamera~\cite{wang2025diffcamera} is constrained to its native $512 \times 512$ training resolution and a fixed aspect ratio, we extract $512 \times 512$ central crops from the DPDD benchmark. This guarantees that both models are evaluated at DiffCamera's optimal setting.

As shown in Table~\ref{tab:supp_diffcamera}, our method consistently outperforms DiffCamera across all full-reference and no-reference metrics. Notably, our approach achieves a significant reduction in LPIPS (0.1285 vs. 0.2833) and DISTS (0.1045 vs. 0.1793). This comprehensive superiority under a strictly controlled baseline confirms our robust generative capacity and highlights the critical advantage of leveraging real-world data during training.

\begin{figure*}[t]
\centering
\includegraphics[width=\linewidth]{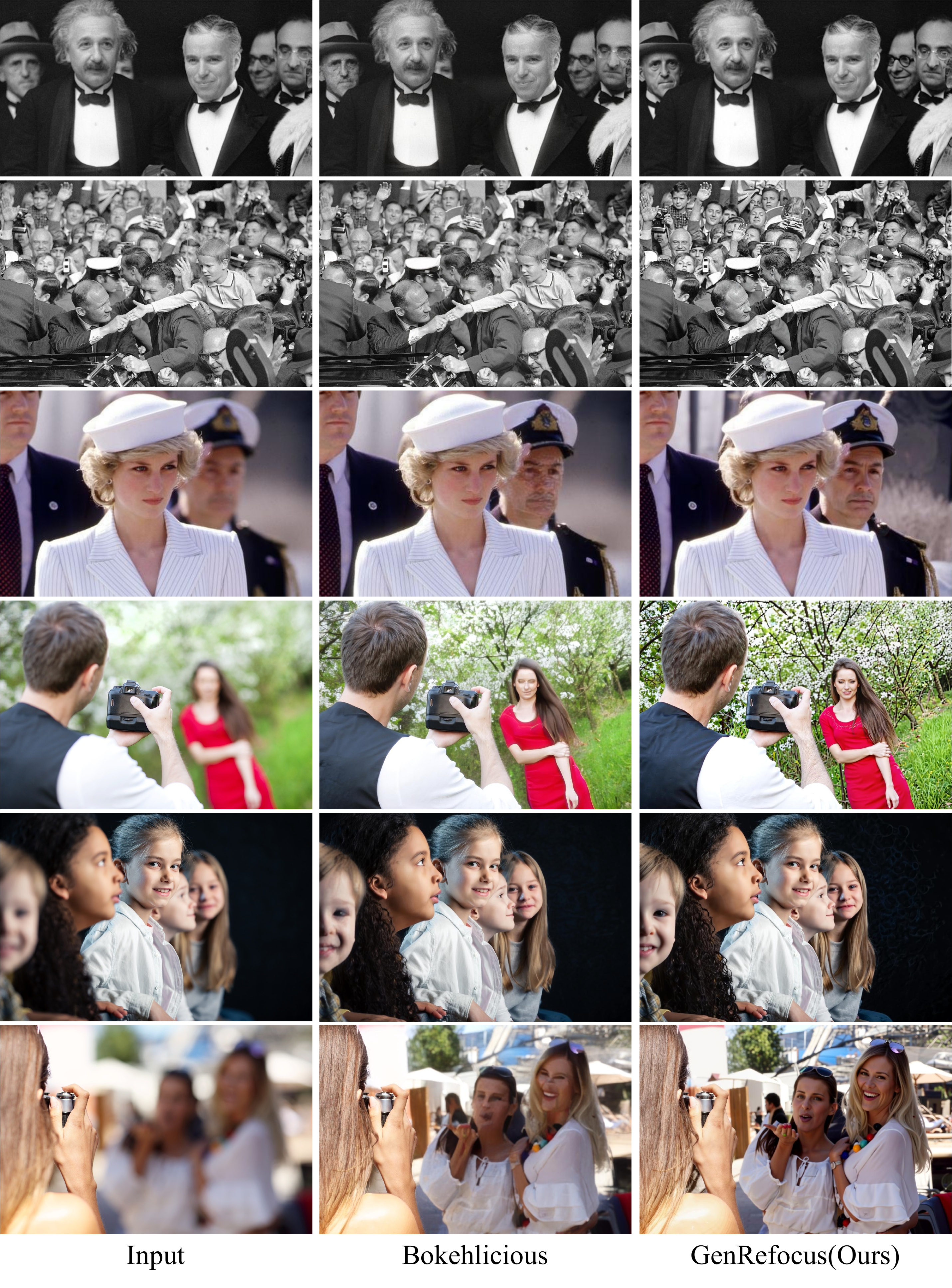}
\caption{
\textbf{Comparison with Bokehlicious~\cite{Bokehlicious}.} Results are shown for historical images (top three rows) and in-the-wild images (bottom three rows). Bokehlicious~\cite{Bokehlicious} either fails to remove the blur entirely or introduces severe structural artifacts. Conversely, our proposed method exhibits strong generalization capabilities, robustly handling diverse blur conditions to synthesize crisp and visually pleasing outputs.
}
\label{fig:Bokehlicious}
\end{figure*}
\section{Qualitative comparison with Bokehlicious}

In the main paper, we state that our model demonstrates greater robustness to diverse input conditions. To further substantiate this claim, we provide additional qualitative comparisons with Bokehlicious~\cite{Bokehlicious} in Fig.~\ref{fig:Bokehlicious}. 

A primary real-world application of defocus deblurring lies in restoring invaluable legacy photographs that can no longer be recaptured, as well as rescuing modern in-the-wild images suffering from accidental misfocus. As shown in the first three rows of Fig.~\ref{fig:Bokehlicious}, we evaluate the models on historical images, including monochrome photographs. Under these conditions, Bokehlicious either yields negligible deblurring effects or introduces severe structural artifacts. Similarly, in the in-the-wild scenarios (the last three rows), Bokehlicious fails to effectively recover the sharp latent images and suffers from visual degradation. 

In contrast, our proposed method successfully restores details across all these highly diverse conditions. From aging black-and-white photos to everyday digital captures, our model consistently produces visually appealing results, thereby validating its strong generalization capabilities and practical applicability.

\begin{figure*}[t]
\centering
\includegraphics[width=\linewidth]{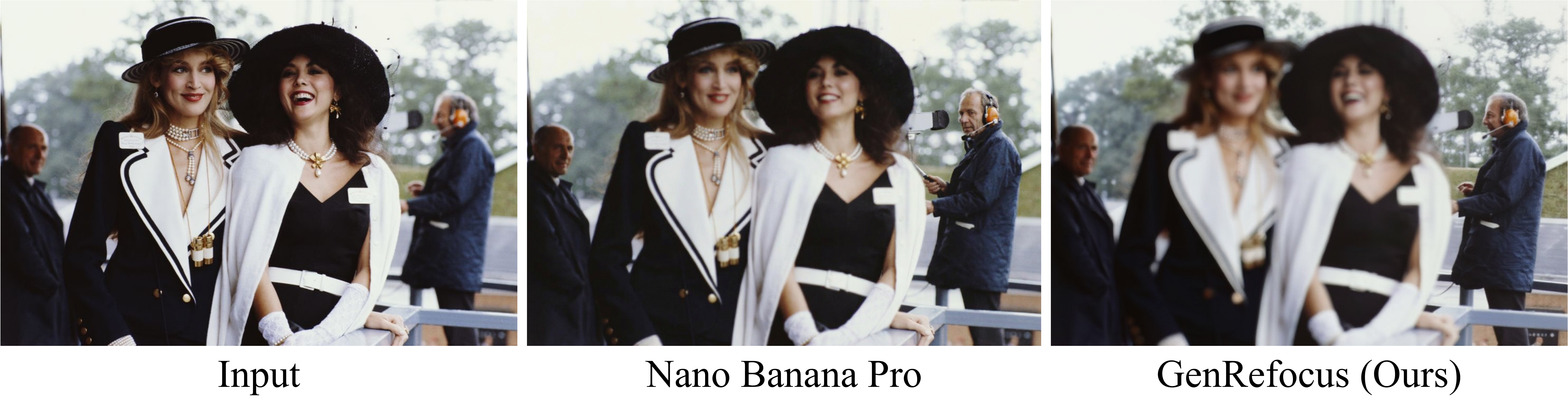}
\caption{
\textbf{Qualitative comparison with VLM model.}
Qualitative refocusing results comparing our Generative Refocusing framework with \emph{Gemini~3 Nano Banana Pro} \cite{googledeepmind_gemini_overview_2025} given prompt "focus on the man on the right". 
}
\label{fig:Banana}
\end{figure*}
\section{Qualitative comparison with VLM model}
We also compare our method with the vision--language model \emph{Nano Banana Pro}~\cite{googledeepmind_gemini_overview_2025}, which explicitly supports text-driven refocusing. For this model, we use the prompt ``focus on the man on the right'' to specify the desired focus point. As shown in Fig.~\ref{fig:Banana}, Nano Banana Pro can change the focus to some extent, but it also alters the facial expressions and appearances of people in both the foreground and the background. In contrast, our method preserves the original content while producing a realistic bokeh effect and accurate refocusing.

\begin{figure*}[t] 
\centering
\includegraphics[width=\linewidth]{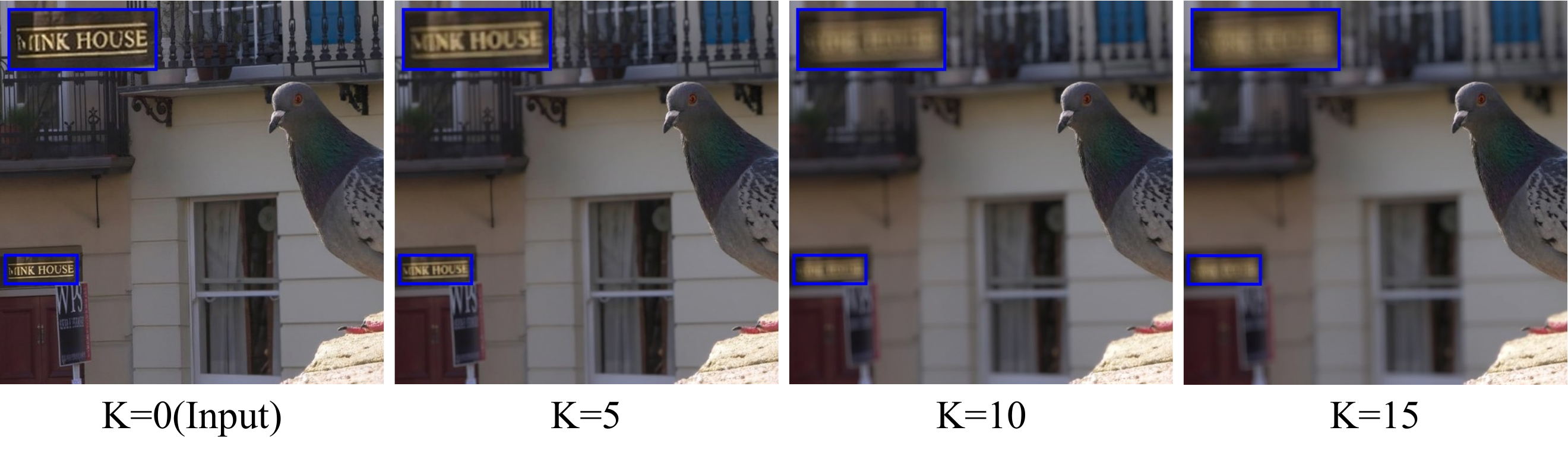}
\caption{
\textbf{Qualitative evaluation of controllability.} 
Refocusing results across different blur levels ($K \in \{0, 5, 10, 15\}$). The \textcolor{blue}{blue} boxes highlight specific text regions to illustrate the defocus effect. As $K$ increases, the blurriness of the scene increases monotonically, demonstrating our model's precise control over the bokeh effect.
}
\label{fig:k_mono}
\end{figure*}

\section{Qualitative Results for Controllability}
In the main paper, we quantitatively demonstrated that our framework outperforms all other diffusion-based baselines in terms of controllability. In this section, we qualitatively evaluate this controllability by examining the influence of the blur parameter $K$ and the defocus map. As shown in figure~\ref{fig:k_mono}, the blurriness of the scene increases monotonically as the given $K$ becomes larger. This monotonic transition effectively proves that our model achieves precise controllability.

\begin{figure*}[t]
\centering
\includegraphics[width=\linewidth]{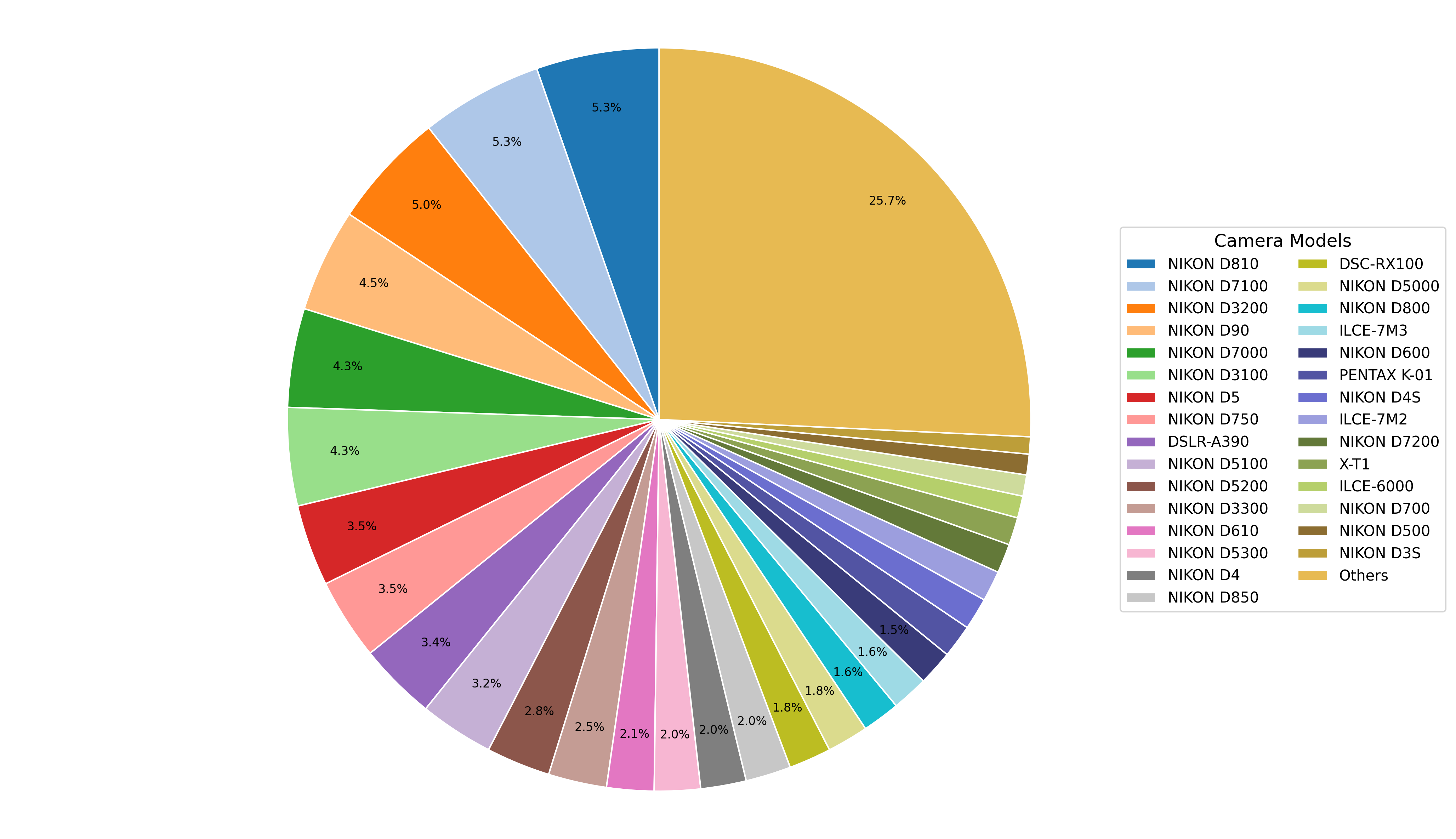}
\caption{
\textbf{Camera model distribution.} 
Percentage of images captured by the top 30 camera models in our dataset. Less frequent models are aggregated into the ``Others'' category.
}
\label{fig:camera}
\end{figure*}

\begin{figure*}[t]
\centering
\includegraphics[width=\linewidth]{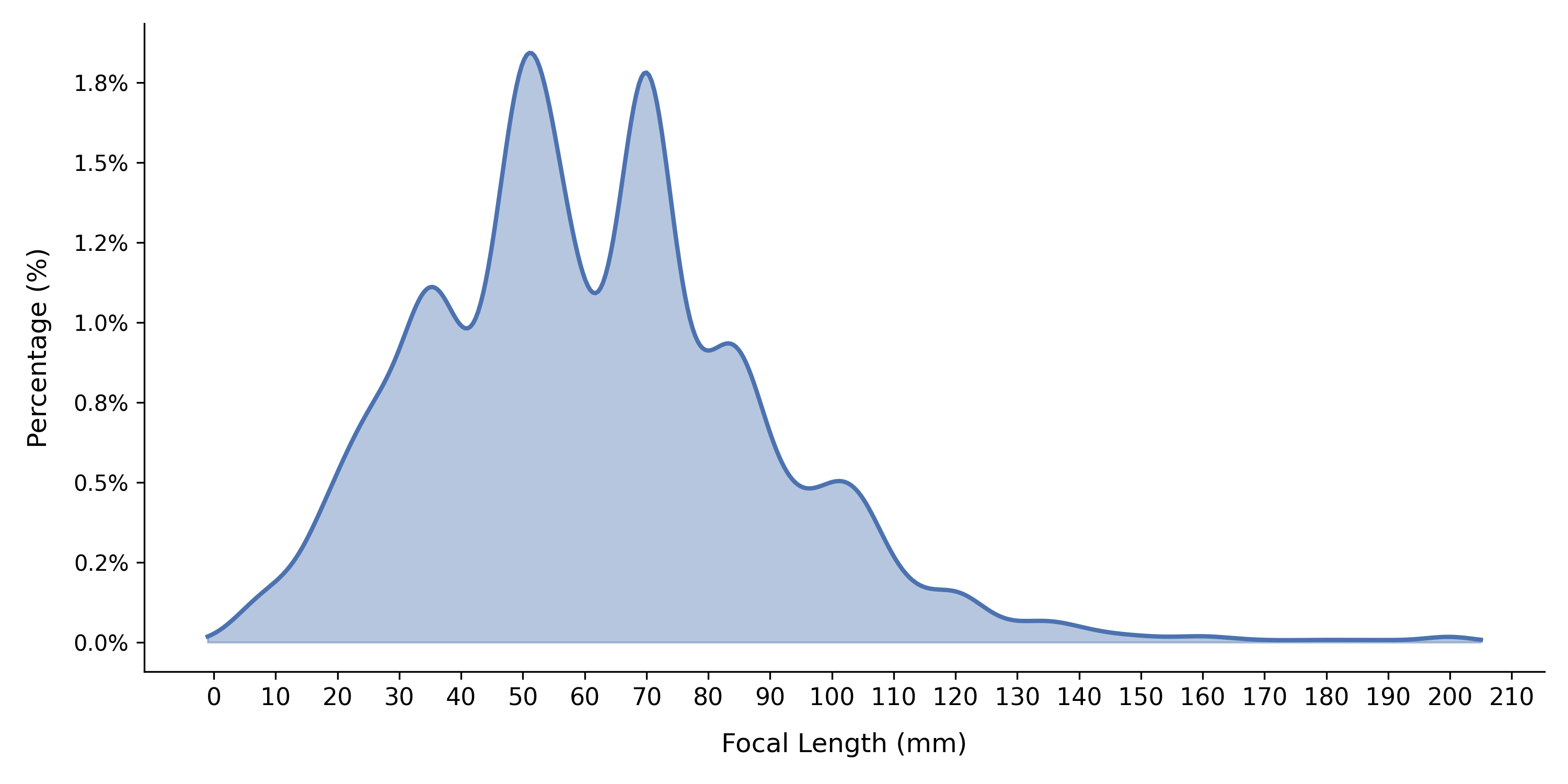}
\caption{
\textbf{Focal length distribution.} 
The percentage distribution of focal lengths across ITW dataset.
}
\label{fig:focal_length}
\end{figure*}

\begin{figure*}[t]
  \centering
  \includegraphics[width=\linewidth]{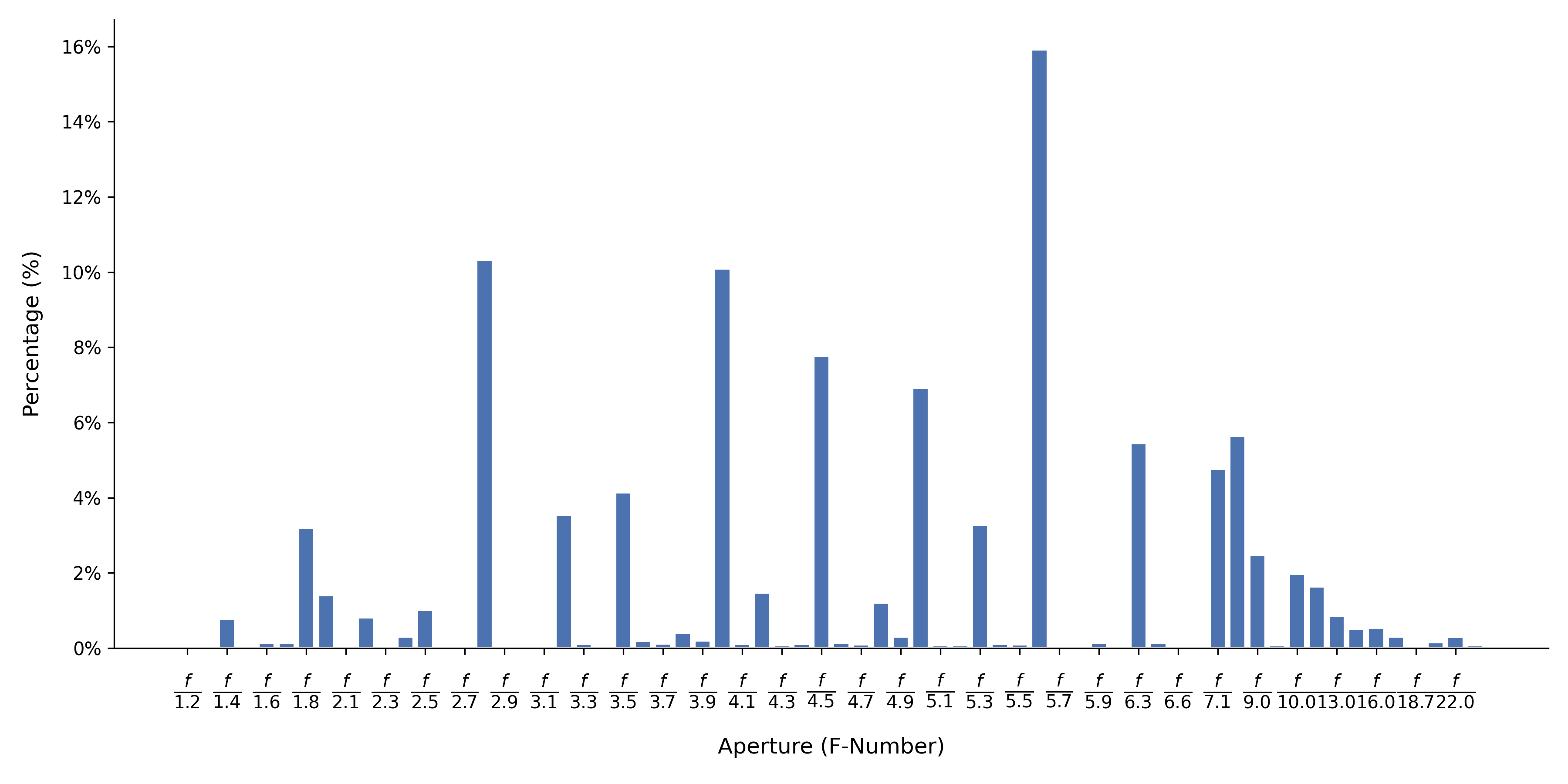}
  \caption{
    \textbf{Aperture (F-number) distribution.} 
    The percentage distribution of aperture settings across ITW dataset. Labels are presented as vertical fractions ($\frac{f}{N}$) for clarity.
  }
  \label{fig:fnumber_dist}
\end{figure*}

\begin{figure*}[t]
\centering
\includegraphics[width=\linewidth]{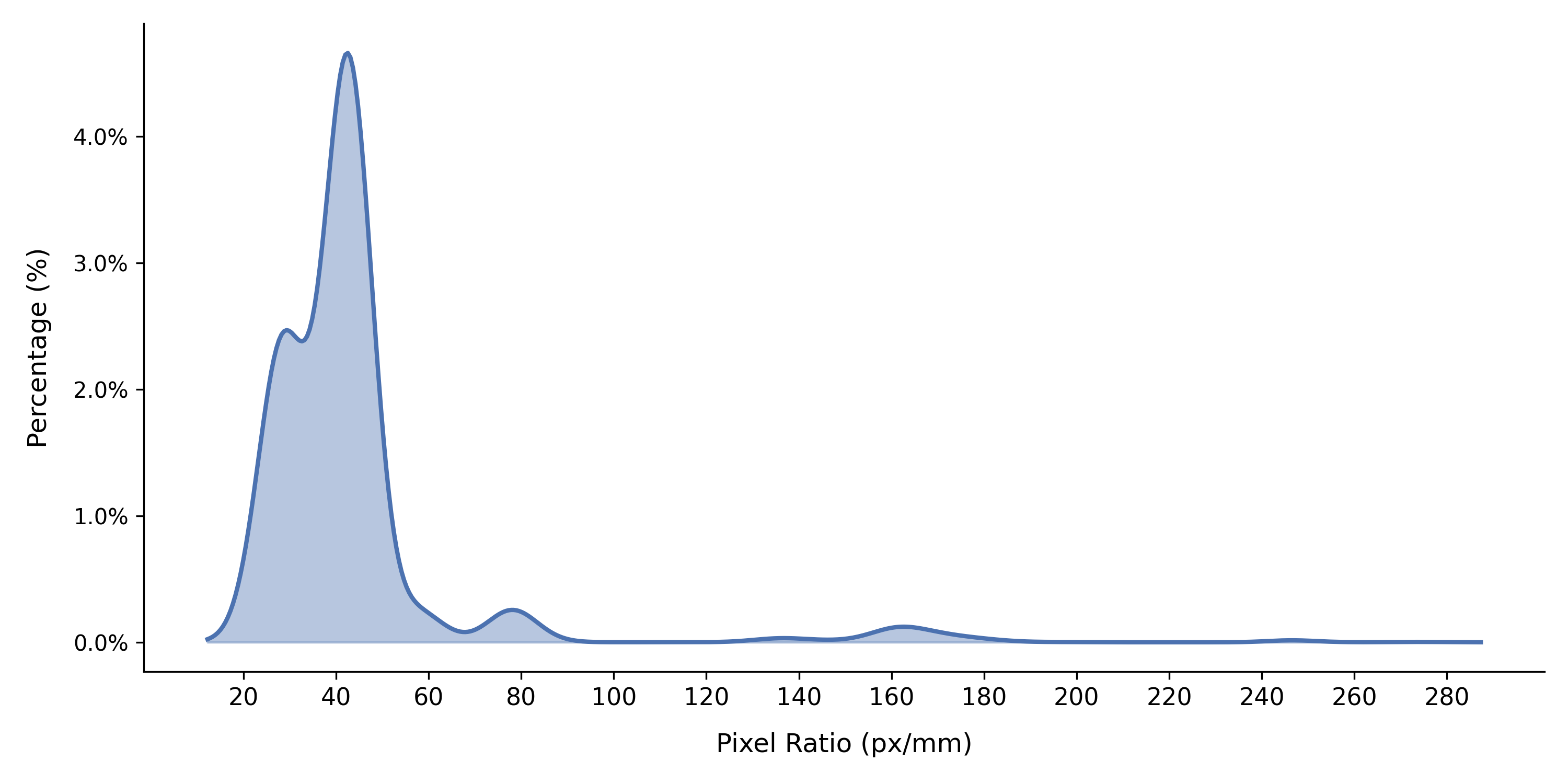}
\caption{
\textbf{Pixel ratio distribution.} 
The percentage distribution of the pixel-to-sensor ratio across ITW dataset. This ratio, defined as the image's largest edge length divided by the physical sensor width ($\text{px}/\text{mm}$).
}
\label{fig:pixel_ratio}
\end{figure*}
\section{ITW Dataset Analysis}
This dataset is utilized to estimate the image bokeh level $K$ via EXIF metadata. We provide further analysis regarding the dataset's diversity. Crucially, all data consists of real optical bokeh images captured directly by digital cameras, rather than artificial defocus synthesized through computational aperture techniques. In addition to the hardware diversity represented by the various camera models (Fig.~\ref{fig:camera}), we analyze the distributions of the key parameters used in our $K$ estimation. Specifically, we provide the distributions for focal length (Fig.~\ref{fig:focal_length}), F-number (Fig.~\ref{fig:fnumber_dist}), and pixel ratio (Fig.~\ref{fig:pixel_ratio}).

\begin{figure}[t]
  \centering
  \includegraphics[width=\linewidth]{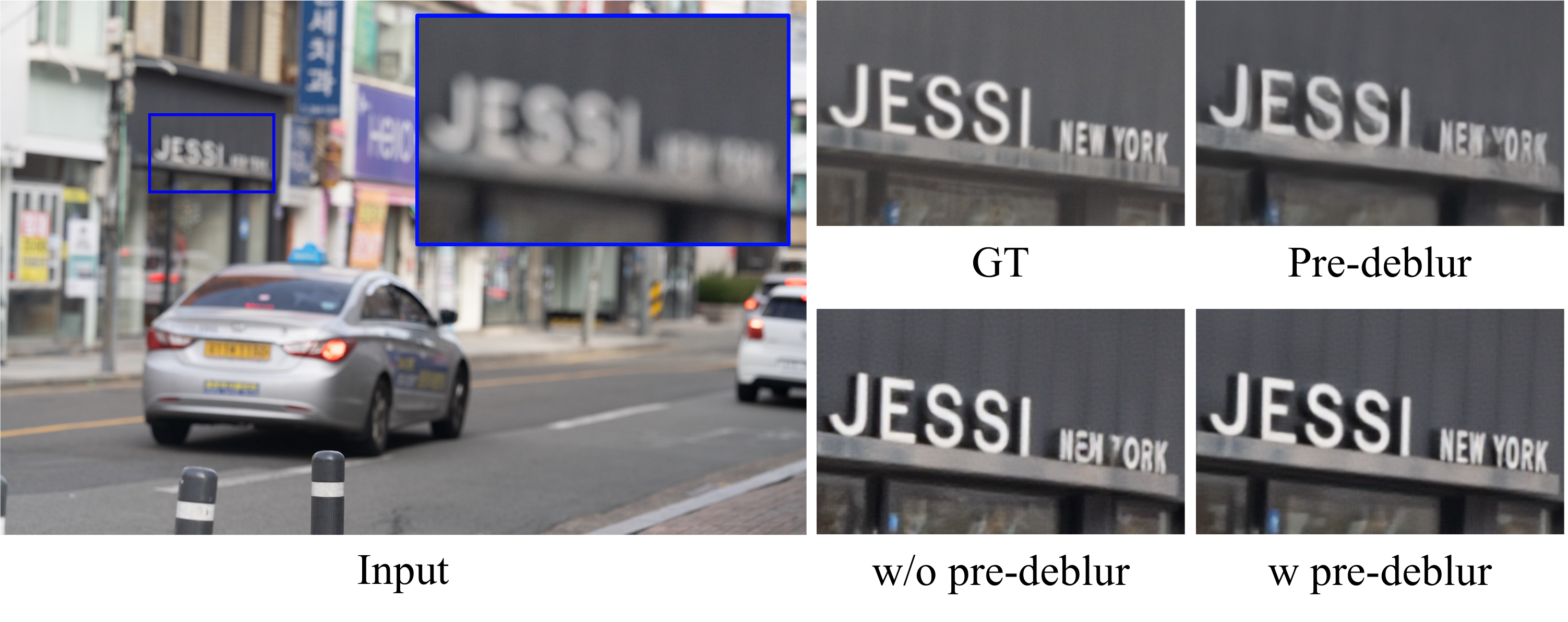}
  \caption{
  \textbf{The effect of the pre-deblur module.} Even when the pre-deblurred image is still distorted and blurry, using it as a condition yields higher-fidelity results (e.g., recovering ``NEW YORK'') than removing this condition, which leads to distorted characters.
  }
  \label{fig:pre_deblur}
\end{figure}

\section{The effect of the pre-deblur module.}
We introduce a variant that conditions DeblurNet on a pre-deblurred estimate. This provides a structural prior to disambiguate severe blur and improve fidelity. As shown in Fig.~\ref{fig:pre_deblur}, even if the pre-deblurred image remains distorted, it helps the model recover fine details (e.g., the text ``NEW YORK'').
\end{document}